\documentclass[acmsmall,nonacm]{acmart}
\usepackage{graphicx}
\usepackage{amsmath}
\usepackage{amsthm}
\usepackage{booktabs}
\usepackage{algorithm}
\usepackage{algorithmic}
\usepackage{amsfonts}
\usepackage{multirow}
\usepackage{makecell}
\usepackage{lscape}
\usepackage{tabularx,rotating}
\AtBeginDocument{%
  \providecommand\BibTeX{{%
    \normalfont B\kern-0.5em{\scshape i\kern-0.25em b}\kern-0.8em\TeX}}}

\setcopyright{acmcopyright}
\copyrightyear{2022}
\acmYear{2022}
\acmDOI{10.1145/1122445.1122456}

\acmJournal{CSUR}
\acmVolume{37}
\acmNumber{4}
\acmArticle{1}
\acmMonth{1}

\begin{document}

\title[A Survey for Curiosity-Driven Learning]{From Psychological Curiosity to Artificial Curiosity: Curiosity-Driven Learning in Artificial Intelligence Tasks}

\author{Chenyu Sun}
\email{chenyu002@e.ntu.edu.sg}
\orcid{0000-0002-2246-7491}
\author{Hangwei Qian}
\email{hangwei.qian@ntu.edu.sg}
\author{Chunyan Miao}
\email{ascymiao@ntu.edu.sg}
\affiliation{%
	\institution{Nanyang Technological University}
	\streetaddress{50 Nanyang Avenue}
	\country{Singapore}
	\postcode{639798}
}


\begin{abstract}
  Psychological curiosity plays a significant role in human intelligence to enhance learning through exploration and information acquisition. In the Artificial Intelligence (AI) community, artificial curiosity provides a natural intrinsic motivation for efficient learning as inspired by human cognitive development; meanwhile, it can bridge the existing gap between AI research and practical application scenarios, such as overfitting, poor generalization, limited training samples, high computational cost, etc. As a result, curiosity-driven learning (CDL) has become increasingly popular, where agents are self-motivated to learn novel knowledge. In this paper, we first present a comprehensive review on the psychological study of curiosity and summarize a unified framework for quantifying curiosity as well as its arousal mechanism. Based on the psychological principle, we further survey the literature of existing CDL methods in the fields of Reinforcement Learning, Recommendation, and Classification, where both advantages and disadvantages as well as future work are discussed. As a result, this work provides fruitful insights for future CDL research and yield possible directions for further improvement.
\end{abstract}

\ccsdesc[500]{General and reference~Surveys and overviews}
\ccsdesc[500]{Computing methodologies~Reinforcement learning}
\ccsdesc[300]{Applied computing~Psychology}
\ccsdesc[100]{Computing methodologies~Cognitive science}
\ccsdesc[100]{Computing methodologies~Intelligent agents}

\keywords{curiosity, curiosity-driven learning, intrinsic reward, collative variable}

\maketitle

\section{Introduction}
\setcitestyle{numbers,sort&compress}
To autonomously produce human-like behaviors in artificial agents, neuroscience has inspired various architectures and algorithms for machine learning (ML), which has obtained phenomenal success in various applications \cite{konar2018artificial,hassabis2017neuroscience,zhu2020transfer}. Apart from leveraging the structure of neurons, the capability of AI systems should also be enhanced with biological and psychological functionality or human-level cognition to inherit the merits of human-level intelligence such as quick adaption, high sample efficiency, and trustworthy interpretation  \cite{wu2013curiosity,oudeyer2009intrinsic,yadav2021human}. As a basic element of cognition, curiosity organically provides an intrinsic motivation driving human beings to explore the world by discovering interesting and useful information \cite{silvia2012curiosity,starr,kidd2015psychology}. Many psychological theories have revealed that exploratory behaviors such as nonstrategic information seeking are stimulated by curiosity internally \cite{berlyne1960conflict,reio2004affect,golman2018information}. As a result, curiosity can affect anxiety, novelty preference, and habituation \cite{hughes1997intrinsic}. The insatiable demand for information can eventually reshape and boost learning, decision-making, as well as healthy development \cite{kidd2015psychology}. In particular, the empirical evaluation indicates a strong association between curiosity and activation in memory regions in the human brain, which benefits the learning of new information \cite{kang2009wick}. 
In recent research, it has become an emerging trend to incorporate curiosity mechanisms in the AI community, where a curiosity-driven learning (CDL) framework can be exploited to enhance the learning performance dramatically \cite{wu2015c,pathak2017curiosity,burda2018large,zhao2019curiositydriven}.

In various AI tasks, there exist many challenges in effective learning and self-conscious evolution, which indicates the gaps between machine-based intelligence and human-level intelligence. In certain types of ML models, agents rely on ``cold" cognition techniques which aggressively mine isolated hierarchical structures \cite{cuzzolin2020knowing}. These methods generally lack a self-driven mechanism for generalization and might suffer from severe overfitting. For example, it is criticized that current approaches of autonomous driving that learn directly from visual observations are insufficient to ensure prediction accuracy as they are simply constructing pure pattern recognition \cite{baker2017rational,rasouli2020attention}. In some practical applications such as navigation and robotic manipulation, training agents to achieve optimal performances requires the direct feedback from the environment to be continuously given; however, it is often expensive or infeasible to acquire them in practice \cite{qiu2016survey,yu2020intrinsic}. Due to the sparse or absent feedback, agents cannot adapt smoothly and their performances are affected substantially. In the big data era, the volume of data to be learned has exploded in diverse domains. A crucial fact is that not all samples are equally important to learn. In other words, selecting the most representative data and extracting meaningful patterns has become the key to reducing computational costs and improving sample efficiency. On the other extreme case, effectively leveraging prior knowledge to obtain good generalization is the most desirable feature in few-shot learning where only limited data is available \cite{wang2020generalizing,qian2021latent}. Although machines have exhibited advantages of fast processing speed and statistical modeling, they lack in-detailed explainability during human-machine interaction \cite{holzinger2018machine}. More specifically, users cannot interpret the model output intuitively, which hinders the application of the ML methods. In the applications such as recommendations and gaming, where extensive human psychological behavior simulations are carried out, machines are expected to closely model and describe the cognitive processing of humans to make human-like decisions. Without a deep understanding of such mechanism, the recommendations or interactions made by AI would be much deviated from the actual human behavior, resulting in unsatisfying user experiences and engagement.  

To address the above-mentioned issues and enable agents to learn actively and efficiently in the most natural way through intrinsic motivation, researchers have started to incorporate artificial curiosity mechanisms into various AI applications \cite{jordan2015machine,togelius2008experiment,schmidhuber2006developmental}.
It essentially strives to combine the merits of intrinsic motivation and encourage spontaneous exploration. Furthermore, different types of curiosity-driven learning (CDL) or curiosity-based learning algorithms have been proposed to perform on the classical 
AI tasks such as classification \cite{stein2017toward}, recommendation \cite{abbas2019one}, optimization \cite{schaul2011curiosity}, and reinforcement learning \cite{schmidhuber1991possibility}, where the ultimate objective is to improve the learning efficiency and potentially enable intelligent agents to learn in a human-like way. For example, when agents actively select and learn the interesting data that they are curious about, it substantially reduces training steps, realizes the self-supervised evolution \cite{pathak2019selfsupervised}, and avoids costly labeling while maintaining similar accuracy \cite{Wang_2017}. 
In the field of reinforcement learning, curiosity is quantified as intrinsic rewards contributing to the learning process. As a result, agents are encouraged towards novel states \cite{10.1145/1143844.1143932,zhao2019curiositydriven,10.5555/2034396.2034466} or perform actions to explore highly uncertain areas based on their existing knowledge about the environment \cite{schmidhuber1991curious,inbook,rezende2015variational,blau2019bayesian}. Driven by curiosity, agents can explore the environment and learn diverse skills that might be even useful in unseen situations, demonstrating the most desirable characteristic of human-like exploratory behaviors.

Although the applications of CDL are vast and could not be exhaustively covered in a single work, \citeauthor{oudeyer2009intrinsic}~\cite{oudeyer2009intrinsic} presented a typology of computational intrinsic motivation which summarized various approaches of intrinsic motivation in psychology as well as computational models in reinforcement learning. \citeauthor{wu2013curiosity}~\cite{wu2013curiosity} introduced a brief computational framework for measuring curiosity and provided some research areas for future studies. However, these works are outdated as more CDL methods have emerged and become increasingly popular since then. In the domain of reinforcement learning, the recent success of leveraging intrinsic rewards (artificial curiosity) to encourage efficient exploration is reviewed and discussed \cite{aubret2019survey,yang2021exploration}. 
Nevertheless, these works lack a comprehensive framework and nuanced understanding of psychological curiosity to differentiate and utilize different types of artificial curiosity fundamentally. In addition, they did not provide insights for formulating other possible artificial curiosity or extending to other domains of AI. 

To the best of our knowledge, our paper is the first comprehensive survey which reviews from the psychological curiosity theory to curiosity-driven structures and algorithms in Reinforcement Learning (RL), Recommendation and Classification. In particular, we aim to present a unified framework and address the key ideas in these major AI tasks with their competing advantages and shortcoming as well as insights for future research. The rest of the paper is organized as follows: Section \ref{psychology} presents the mechanism of curiosity from the psychological perspective, which provides an important insight and guideline for proposing artificial curiosity with an arousal mechanism. Section \ref{RL}, Section \ref{Recommendation} and Section \ref{classification} respectively survey the curiosity-driven structures and algorithms applied to RL, Recommendation, and Classification tasks, from motivations to merits and limitations. Finally, Section \ref{conclusion} discusses some future research directions and challenging issues for curiosity-driven learning.

\section{Psychological Theories of Curiosity}
\label{psychology}
\subsection{Development of Psychological Curiosity Theories}
Over the past few decades, extensive research and empirical analysis have been carried out on biological mechanisms of curiosity in behavioral, neurobiological, and cognitive studies \cite{dobrynin2009physical,kashdan2009curiosity,kidd2015psychology,jepma2012neural}. 
Nevertheless, these studies are yet inconclusive and often contradictory. In the first half of the 20th century, curiosity was classified as an emotion closely related to fear \cite{hall1903curiosity}. At that time, drive theories became popular, whereas curiosity might be associated with an unpleasant feeling of deprivation and can be reduced through a series of exploratory behaviors \cite{maslow1943theory,berlyne1950novelty}. Thereafter, the first complete theoretical framework based on drives was presented by K. Lorenz \cite{lorenz1981exploratory}, which provided the research basis for further studies invariably. However, it was inconclusive about whether curiosity is a primary or secondary (derived from other drives) drive and homeostatic or stimulus-induced drive \cite{loewenstein1994psychology}. In the second half of the 20th century, incongruity theories were developed independently by \citeauthor{hebb1946nature}~\cite{hebb1946nature} and \citeauthor{piaget2003psychology}~\cite{piaget2003psychology}, where curiosity is believed to be triggered by the violation of expectation based on existing knowledge. The relationship between the stimulus and curiosity can be described by an inverted U-shape. The Competence and Intrinsic Motivation theory \cite{white1959motivation} (extended in \cite{deci2010intrinsic}) overthrew both drive theory and incongruity theory that curiosity is a ``competence” motivation triggered by the desire to master the environment. Yet, this theory did not present a comprehensive framework of curiosity. 

In the 1950s, most research of curiosity focused on psychological underpinnings \cite{dember1957analysis}. Later, researchers began to quantify curiosity and evaluate its dimensionality. In particular, \citeauthor{berlyne1960conflict}~\cite{berlyne1960conflict} first proposed a systematic measurement about cognitive curiosity, where curiosity is a drive-based motive aroused by stimulus and human brains generate intrinsic rewards for learning and information acquisition. He summarized several quantifiable \textbf{collative variables} for curiosity activation aroused in the organic level. Based on the psychological curiosity theory presented by \citeauthor{berlyne1960conflict}~\cite{berlyne1960conflict}, \citeauthor{loewenstein1994psychology}~\cite{loewenstein1994psychology} proposed a knowledge gap theory, stating that manageable knowledge gaps could stimulate curiosity by conceptualized novelty, complexity, and ambiguity. Meanwhile, the optimal stimulation model hypothesized both positive and negative affective states on curiosity activation \cite{starr}, where a consistent hypothesis is proposed in the I/D model \cite{lievens2021killing}. Recently, it was shown that the intrinsic reward processing could substantially induce curiosity that is crucial and beneficial for learning and decision-making at both individual and social levels \cite{litman2008interest,jepma2012neural,silvia2012curiosity,kidd2015psychology,wagstaff2021measures}. In 2009, the psychological curiosity theory proposed by \citeauthor{berlyne1960conflict}~\cite{berlyne1960conflict} was validated by conducting experiments on how the monkey’s dopaminergic system responded to rewards \cite{bromberg2009midbrain} and later it was shown that the intrinsic reward processing could substantially induce curiosity in humans \cite{kang2009wick,jepma2012neural,wiggin2019curiosity}. These findings have highlighted the significance of psychological curiosity and revealed the functionality of intrinsic rewards biologically. 

In this paper, we mainly focus on the theory proposed by \citeauthor{berlyne1960conflict}~\cite{berlyne1960conflict} as a quantitative way to measure curiosity. Meanwhile, the knowledge gap theory \cite{loewenstein1994psychology} and the optimal stimulation model \cite{starr} are leveraged to explain the curiosity arousal and preference. More details are provided in Section \ref{framework}.

\subsection{Effects of Curiosity}
Despite the differences in the aforementioned theories and findings, we can almost identify and conclude the desirable effects of psychological curiosity in the development of human intelligence, which motivates to enable AI agents to learn and perform in a similar mechanism. 

Firstly, curiosity can encourage efficient information acquisition and exploration. As a result, attentional resources can be directed to enhance memory for certain information and it benefits the learning process to become more efficient. The empirical works have shown that curiosity could lead people to engage in more elaborative encoding, which leads to memory enhancement \cite{kang2009wick,mcgillivray2015thirst}. Similarly, this intrinsic desire for information could characterize the capability of infants to implicitly allocate their attention towards certain information features with intermediate rates of information absorption \cite{kidd2012goldilocks,kidd2015psychology}, helping to optimize their knowledge acquisition process during exploration. 

Secondly, curiosity can improve human learning capability and promote efficiency. It has been found that individuals generally experience curiosity by encountering the information that is novel or unexpected, and such interest can facilitate and enhance learning and memory \cite{fandakova2021states}. At the organic level, this phenomenon can be explained by the enhanced hippocampal activity that is regulated by dopamine and norepinephrine when experiencing stimulus-evoked curiosity \cite{bunzeck2012contextual,oudeyer2016intrinsic}.
In the education literature, it is encouraged to motivate students to learn in a curious way so that they can actively participate in the learning process while passive enforcement may result in low learning efficiency \cite{johnson2008active}. The empirical evidence has also indicated the association between curiosity and better learning outcomes \cite{gruber2014states,kang2009wick,stahl2015observing}. 

Thirdly, curiosity can be beneficial for building up social bonds and avoiding group conflict \cite{gino2018business}. In particular, social interactions are more ambiguous and complex, which provides more opportunities for self-expansion and requires participants to cope with uncertainty under the mechanism of curiosity \cite{kashdan2009curiosity}. Moreover, curious people tend to be more responsive and self-disclosure to seek out close and collaborative relationships \cite{koo2010interactional,lievens2021killing,wagstaff2021measures}. As a response, the interaction partners are typically willing to explore possible growth opportunities through engaging, learning, and even taking risks \cite{lievens2021killing}. As a result, being curious at social levels is more likely to shape positive and healthy social relationships with meaningful interactions. 

Fourthly, curiosity is highly correlated to human emotions and feelings, such as pleasure, satisfaction, uncertainty, deprivation and frustration. The I/D model proposed by \citeauthor{litman2008interest}~\cite{litman2008interest} considers curiosity as a combination of interest and deprivation, where learning something novel can be entertaining and aesthetically pleasing while eliminating a specific knowledge gap can become increasingly uncomfortable and bothersome. The empirical evidence has implied that curiosity can elicit more positive emotions, which subsequently mediate a positively biased evaluation towards the advertised products \cite{daume2020curiosity}.

Lastly, curiosity has a positive correlation with creativity and innovation. In particular, curiosity promotes information acquisition, analysis and problem identification, which are the key steps for initializing the creativity process \cite{mumford2017creative}. In addition, as curiosity can foster pleasure with positive feelings, the creative responses are generated favorably, including new ideas and behaviors different from prior experiences \cite{schutte2020meta}. The empirical results indicate that this process is mediated through a mechanism of idea linking, where early ideas can sequentially foster subsequent ideas \cite{hagtvedt2019curiosity}. More broadly, the association between curiosity and creativity can subsequently facilitate the development of skills and self-expansion.

\subsection{Framework for Quantitative Measurement with Arousal Mechanism}
\label{framework}
Given the promising and desirable merits of psychological curiosity that promote human learning and intellectual development, a direct intuition is to propose a similar framework for machine-based intelligence. However, it is changeling to quantitatively measure the degree of curiosity in an integrative framework. As suggested in the survey work \cite{wu2013curiosity}, the psychological curiosity theory presented by \citeauthor{berlyne1960conflict}~\cite{berlyne1960conflict} can be introduced as the baseline of measuring artificial curiosity computationally, by determining the stimulus (received from environments) intensity with one or more collative variables. In particular, different levels of stimulus intensity correspond to different levels of curiosity with preferences. In this section, we systematically introduce different collative variables that can be utilized to examine stimulus intensity, which serves to formulate different quantitative ways for measuring artificial curiosity in a unified framework. To describe the arousal principle, the manageable knowledge gap theory \cite{loewenstein1994psychology} as well as the optimal stimulation/dual-process theory \cite{starr} are introduced. Inspired by the arousal principle of psychological curiosity, we also provide some insights for efficient learning rules as well as optimization methods given the measured curiosity.

\subsubsection{Key Assumptions}
In order to integrate the curiosity mechanism into Artificial Intelligence seamlessly without loss of generosity, we assume that 
\begin{itemize}
	\item Curiosity is a primary drive while it is not derived from other basic drives such as hunger and thirst for simplicity \cite{kidd2015psychology}. 
	\item Curiosity is a stimulus-induced drive where stimulus intensity could be measured through a combination of collative variables \cite{berlyne1960conflict}.  
	\item Curiosity can activate both positive feelings and negative deprivation. The net effects depend on the degree of curiosity \cite{starr}.
\end{itemize}

\subsubsection{Collative Variables}
\label{collative variable}
A quantitative way of measuring curiosity is through one or multiple collative variables, given a stimulus from environments. In particular, collative variables are a collation of information from various sources, which detect the similarities and dissimilarities towards the environment based on the prior knowledge or expectation in the related fields \cite{berlyne1978curiosity,silvia2005cognitive}. The main collative variables are defined as follows:

\textbf{Novelty} refers to the perceived ``new” stimulus or information (experiences), which can be classified into different types. 
Its degree is inversely related to 1) how often the similar stimulus has been encountered before, 2) how recently the stimulus has been experienced, and 3) how similar the stimulus is compared to the previously encountered ones.

\textbf{Change} represents the movement when a stimulus is acting on receptors. It can be measured by the differences between two time steps. It should be noted that change can be utilized as a supplementary variable to novelty.

\textbf{Surprisingness} arises when the stimulus occurs differently from the expectation. In other words, $Y$ is expected to occur given a stimulus $X$; however, when $Z$ is experienced in contrast to $Y$, it will result in surprisingness. Surprisingness can also be utilized as a supplementary variable to novelty.

\textbf{Incongruity} occurs when the stimulus induces an expectation that does not occur ultimately. In other words, $Y$ is expected not to occur given a stimulus $X$, so incongruity presents when $Y$ is observed. 
Incongruity can also be utilized as a supplementary variable to novelty.

\textbf{Complexity} refers to heterogeneous elements and describes how they are distributed spatially and temporally in a stimulus. The degree of complexity can be measured by 1) the number of distinguishable units in a stimulus, 2) the dissimilarity among units, and 3) the degree of responses provided by several elements as a whole (cohesion).

\textbf{Uncertainty} refers to the situation where no deterministic response could be selected in response to a stimulus. The degree of uncertainty can be quantitatively measured through various forms of entropy according to information theory, by recognizing (classifying) the received stimulus or the response into different categories. 
In this measure, uncertainty is high when the received stimulus or the response cannot be confidently recognized. 

\textbf{Conflict} refers to the situation of incompatible responses that could not be simultaneously present. The degree of conflict can be measured through 1) the nearness to equality in the strength of the incompatible responses, 2) the absolute strength of the competing incompatible responses, 3) the number of incompatible responses, and 4) the degree of incompatibility between incompatible responses.

\subsubsection{Arousal Mechanism}
As we have assumed that curiosity is stimulus-evoked, it is important to measure the stimulus intensity quantitatively, by one or multiple collative variables introduced in Section~\ref{collative variable}. Then, different levels of stimulus intensity can arouse different levels of curiosity as well as exploratory behaviors. According to \citeauthor{berlyne1960conflict}’s hypothesis~\cite{berlyne1960conflict}, the relation between attractiveness (curiosity) and stimulus intensity is an inverted U-shape. When the perceived stimulus gives low intensity, it would cause boredom and motivate people towards diverse exploration by aimlessly searching stimulus with greater intensity. When the perceived stimulus gives high intensity, the person would engage in specific exploration to reduce the aroused curiosity and its associated negative feelings. Finally, the ultimate arousal would be stabilized toward the desired moderate level. 

Based on this mechanism, \citeauthor{wu2013curiosity}~\cite{wu2013curiosity} proposed two principles namely Avoidance of Boredom and Avoidance of Anxiety, which are consistent and coupled with the Wundt curve \cite{berlyne1960conflict}. However, it only explained the net curiosity preference based on different levels of stimulus intensity, but did not comprehend the full arousal mechanism. Therefore, we provide more insights with two additional systems (a reward system and an anxiety aversion system) on the bell-shaped curve of the curiosity preference function as shown in Figure\ref{InvertedU}. Our extension in this section is based on the aid of manageable knowledge gap theory \cite{loewenstein1994psychology} as well as the optimal stimulation/dual-process theory of exploratory behavior \cite{starr}.

In particular, the knowledge gap theory \cite{loewenstein1994psychology} intuitively explains that humans prefer moderate levels of stimulus intensity as measured by one or multiple collative variables. Moreover, the intensity of a person's curiosity towards a specific item or information depends on his/her ability to resolve the information gap. To stimulate an optimal level of curiosity, learners should be aware of manageable gaps in their knowledge, where extremely large gaps could substantially discourage learning. Similarly, learners are apathetic to the challenge when gaps are too small. 

Meanwhile, \citeauthor{starr}~\cite{starr} proposed an optimal stimulation model illustrating the effect of stimulus characteristics of activating both positive and negative affective states. They emphasized that the purpose of exploration is to induce and increase the pleasantness associated with curiosity. In order to arouse and maintain a sense of curiosity, organisms are excited to seek unusual and novel stimulus. However, since stimulus may also indicate a potential danger, curiosity and exploration might be suppressed when a certain degree of threat is encountered. They concluded that collative variables are the main source of curiosity and the optimal level of curiosity is simultaneously influenced by two systems: the reward system (pleasant states of curiosity) and anxiety aversion system as shown in Figure~\ref{InvertedU}. The following findings can be observed in Figure~\ref{InvertedU}:
\begin{figure}
	\centering
	\includegraphics[width=0.45\textwidth]{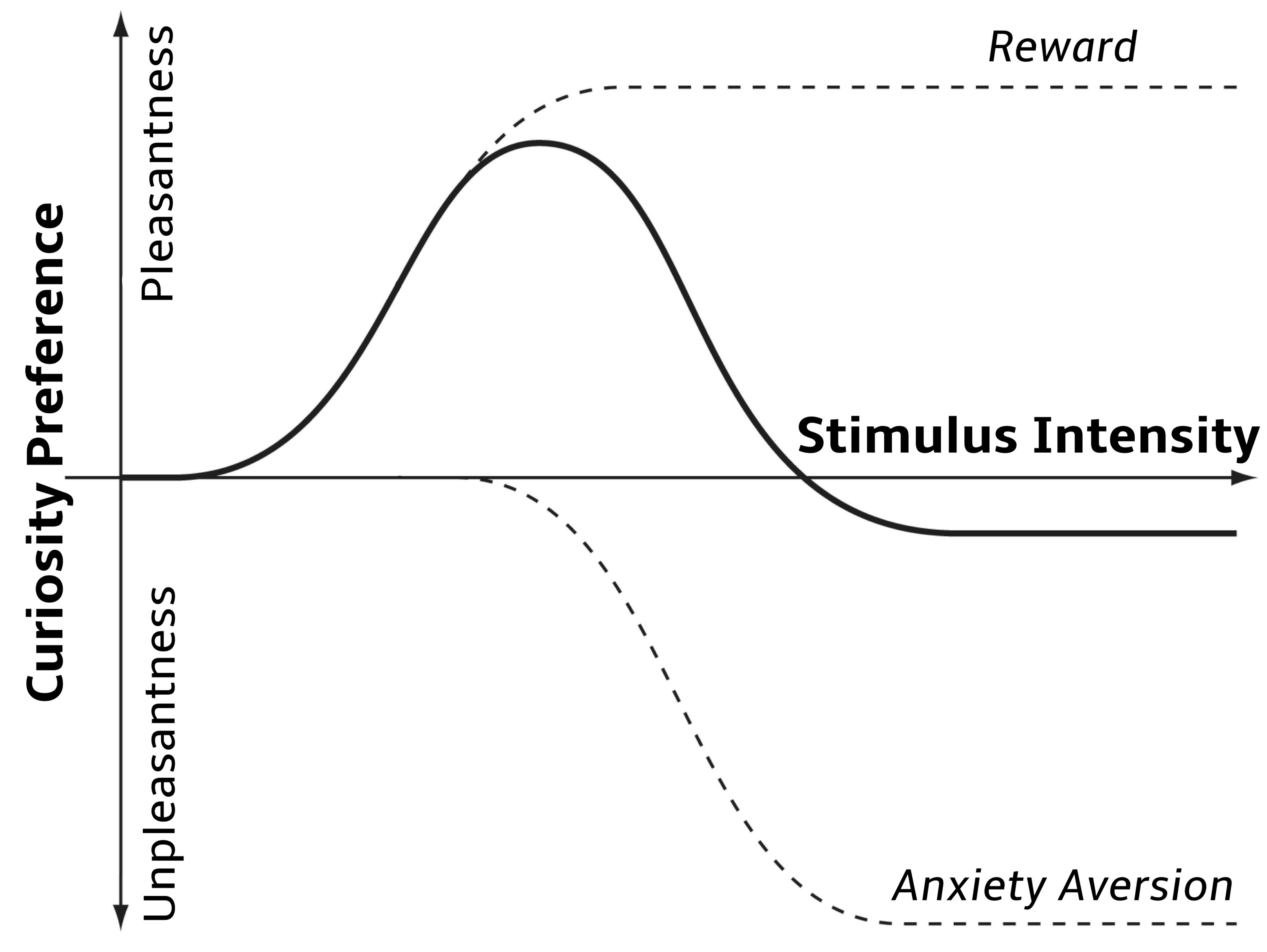}
	\caption{Inverted U-shaped Curiosity Preference Function: the aroused curiosity first increases to the highest then decreases rapidly as the increase of stimulus intensity. Beyond some threshold, the unpleasant feeling will drive the individual away from the stimulus, indicating a flat negative curiosity preference. Thus, the optimal stimulus intensity would be located at some intermediate level and any downward or upward deviation from this optimum would result in interest reducing or aversion.}
	\label{InvertedU}
\end{figure}
\begin{itemize}
	\item Both pleasantness (positive reward) and anxiety aversion (negative reward) increase with the collative stimulus intensity, where the reward system stimulates the exploratory behaviors while the anxiety aversion system motivates the stimulus avoidance.
	\item By aggregating the reward and anxiety aversion systems, the curiosity preference shows a bell-shaped curve w.r.t. stimulation intensity, resulting from different thresholds, growth rates and asymptotic levels of the reward and anxiety aversion systems dynamically. The overall effect is consistent with the Wundt's curve as well as Berlyne’s hypothesis.
	\item With the increase of collative stimulus intensity in an early stage, the reward system dominates the anxiety aversion system, leading to an increase of curiosity feelings and arousal up to some optimal level of effectiveness.
	\item With the increase of collative stimulus intensity in a later stage, the anxiety aversion system rapidly dominates the reward system, resulting in the increase of unpleasantness and its associated behaviors such as avoidance and withdrawal.
	\item Asymptotically, the anxiety aversion system is superior to the reward system, resulting in net unpleasantness. In other words, the cap of the reward system is lower than that of the anxiety aversion system. Beyond some threshold of stimulus intensity, the net unpleasantness will eventually cause the agent to keep away form the stimulus.
	\item As the sensitivity towards reward and anxiety aversion is not the same among different individuals, this bell-shaped curiosity preference function as well as the optimum are not necessarily unique when describing different individuals.
\end{itemize}

\subsection{Discussion}
In summary, a brief review of various psychological theories about curiosity is introduced in this section. In particular, we first focus on the curiosity measurement principle proposed by \citeauthor{berlyne1960conflict}~\cite{berlyne1960conflict}, where several collative variables are introduced and well defined to quantify the stimulus intensity. Subsequently, to determine the level of aroused curiosity, the knowledge gap theory as well as the dual process model are explained, which are extensions of the prior work for enlightening a more comprehensive and solid computational framework of artificial curiosity. The reward and aversion systems explain clearly on the bell-shaped curve between stimulus intensity and curiosity level. An increase of stimulus intensity will firstly activate a reward system and then an antagonistic aversion system. Certainly, this framework systematically illustrates how curiosity is aroused from the psychological perspective, which can also serve as a reference for formulating artificial curiosity in a diversified way.
Yet, there indeed exist some limitations in this framework to some extent. For example, 
some ecological properties such as thirst, hunger and fear have been shown to be capable of affecting curiosity as well \cite{zillmann2018attribution}, which is also an important determinant for exploratory behaviors. For simplicity, we did not consider these properties and their impacts in this paper. 
In the following sections, we further survey the literature of curiosity-driven learning methods in the domains of Reinforcement Learning, Recommendation, and Classification based on the psychological principle as discussed in this section. In particular, we will focus on both their comparative advantages and weakness as well as provide some insights for future work.

\section{CDL in Reinforcement Learning}
\label{RL}
\begin{figure}
	\centering
	\includegraphics[width=0.6\textwidth]{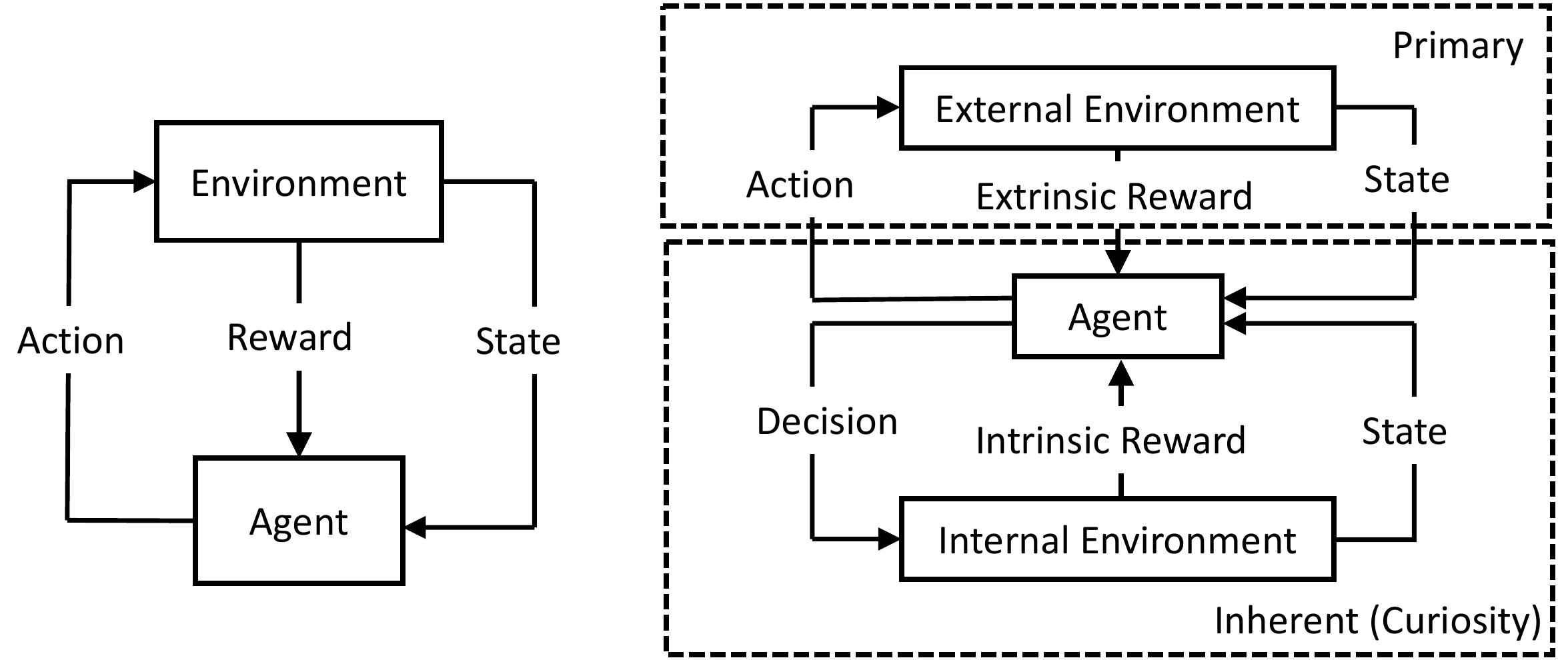}
	\caption{Agent–Environment Interactions in RL
	}
	\label{RLint}
\end{figure}
In the domain of reinforcement learning (RL), Figure~\ref{RLint}~(left) illustrates how an agent interact with environments to perform the action $a_t$ given the state $s_t$ at each step $t$, by receiving the reward $r_t$ and observing the next state $s_{t+1}$ with a transition model $p(s_{t+1}|s_t, a_t)$. In particular, the Markov decision process (MDP) assumes that the transition model $p(s_{t+1}|s_t, a_t)$ depends only on the current state $s_t$ and action $a_t$ rather than the historical actions or states and the agent should take the action $a_t$ only based on the current state $s_t$. For the received reward $r_t$, it can be factored into two components: intrinsic rewards $r_t^i$ and extrinsic rewards $r_t^e$. As illustrated in Figure\ref{RLint}~(right), the extrinsic rewards are the primary feedback provided by the external environment directly while the intrinsic rewards are the interest (curiosity) level that the agent precepts from its internal environment inherently \cite{barto2004intrinsically}. Moreover, the actions and states can be either discrete or continuous while the states can be high-dimensional visual observations $o_t$ such as raw pixels, which increases the complexity of the underlying RL problems to some extent. Nevertheless, the ultimate goal is to learn an optimal policy $\pi$ to complete target tasks by maximizing the expected rewards from the actions taken in the environment, i.e. $\mathbb{E}\left[{\sum_t r_t}\right]$. It has been recently shown that reward maximization provides a fundamental basis for developing broadly capable intelligence and skills such as perception, language, generalization and imitation \cite{silver2021reward}.

There are plenty of interesting yet challenging problems that can be addressed through curiosity-driven learning (CDL) techniques in the RL research community. For example,
conventional RL algorithms can perform well when $r_t^e$ are explicitly and continuously given. Intelligence and its associated abilities can implicitly arise through reward maximization when an agent acts in its environment \cite{silver2021reward}. However, in the ubiquitous challenge of absent or sparse $r_t^e$, $r_t^i$ plays an important role in determining where to explore next \cite{chentanez2004intrinsically}. In this sense, the intrinsic reward $r_t^i$ can be treated as artificial curiosity, which motivates agents to actively explore the most interesting states and obtain a desirable generalization capability \cite{bellemare2016unifying,pathak2017curiosity,houthooft2016vime,berseth2020smirl,li2021mural}.
Moreover, training an RL model typically requires extensively collecting trajectories $\tau_t = (s_t, a_t, r_t, s_{t+1})$ into the replay buffer with GPU resources, where the imbalanced data problem may hinder the convergence of learned policies or result in sub-optimal policies. Therefore, it is desired to employ artificial curiosity and encourage agents to focus more on the under-explored states by prioritizing the experience replay  \cite{schaul2016prioritized,kapturowski2018recurrent,zhao2019curiosity,brittain2019prioritized,jiang2021prioritized}. 
Another challenge in RL is sample efficiency, especially when dealing with high-dimensional observation inputs as such raw pixels \cite{malik2021sample,yarats2021improving,pmlr-v139-malik21c}. Hence, efficiently extracting meaningful feature representation with the limited amount of collected trajectories is the main motivation of utilizing CDL methods \cite{nguyen2021sample}. 

In this section, we survey the RL literature that incorporates different types of artificial curiosity based on the quantitative measurement framework as introduced in Section \ref{framework}. In particular, various \textbf{collative variables} are structurally utilized to evoke the artificial curiosity with the arousal mechanism. Subsequently, the proposed curiosity intrinsically encourages agents to balance the exploration and exploitation through a diverse range of intrinsic rewards or experience replay prioritization. As a result, agents can quickly learn an optimal policy to perform complex tasks under different settings. However, as the formulation of artificial curiosity is not unique, we will also discuss the merits as well as limitations for each type of curiosity, and provide some insights for future CDL-based RL research.


\subsection{Uncertainty-Based Curiosity}
\label{Uncertainty-Based Reward}
The uncertainty-based curiosity mainly considers the agent's uncertainty towards the under-explored states or the possible actions to take, so it encourages agents to explore the states or take actions such that the uncertainty can be reduced while information gain can be obtained. 
For continuous control tasks where the state-action space grows
exponentially in its dimensionality, heuristic models commonly employ Gaussian control noise \cite{duan2016benchmarking} or $l^2$ neural network prediction \cite{stadie2015incentivizing} as the intrinsic rewards $r^i$ to encourage more active exploration. However, these methods may cause inefficient exploration and random walk behaviors that may require the training time to be exponential in the number of states. To remedy this problem, \citeauthor{houthooft2016vime}~\cite{houthooft2016vime} proposed a Variational Information Maximizing Exploration (VIME) strategy, which makes full use of information gain about the agent’s internal belief towards the environments as an intrinsic learning motivation. In this strategy, curiosity is modeled by the agent's \textit{uncertainty reduction} about the environment dynamics, and is directly employed as the intrinsic reward. Its intuition is to encourage agents to explore the next state that can reduce its uncertainty the most. According to information theory, uncertainty reduction (information gain) refers to the change in mutual information $I(S_{t+1};\Theta | \xi_t, a_t)=H(\Theta|\xi_t,a_t)-H(\Theta| S_{t+1},\xi_t,a_t) $ between the next state $S_{t+1}$ and the transition model parameter $\Theta$, where $H(\cdot)$ represents the entropy and $\xi_t=\{s_1,a_1,s_2,a_2,\cdots,s_t\}$. Explicitly, it can be proved that
\begin{equation}
I(S_{t+1};\Theta | \xi_t, a_t)=\mathbb{E}_{s_{t+1}\sim P(\cdot | \xi_t,a_t)}\left[D_{KL}\left[p(\theta | \xi_t, a_t,s_{t+1})\parallel p(\theta |\xi_t)\right]\right],
\end{equation}
where the Kullback–Leibler (KL) divergence $D_{KL}\left[p(\theta\mid \xi_t, a_t,s_{t+1})\parallel p(\theta\mid\xi_t)\right]$ captures the change in an agents internal belief after observing new data over the dynamics model. 
Given deterministic next state $S_{t+1}=s_{t+1}$, the intrinsic reward at time step $t$ can be defined by
\begin{equation}
r_t^i=\eta D_{KL}\left[p(\theta\mid \xi_t, a_t,s_{t+1})\parallel p(\theta\mid\xi_t)\right],
\label{uncertainty-based reward}
\end{equation}
where $\eta \in \mathbb{R^+}$ is a hyper-parameter for scaling.

However, the most challenging problem is that the posterior distribution $p(\theta\mid \xi_t, a_t,s_{t+1})$ is generally intractable in most cases. Therefore, the variational interference is employed to approximate it with $q(\theta;\psi)$ using a Bayesian neural network, where $\psi$ is the parameter and it is optimized by minimizing $D_{KL}\left[q(\theta;\psi)\parallel p(\theta\mid\xi_t,a_t,s_{t+1})\right]$. As a result, 
the intrinsic reward in Eq.~(\ref{uncertainty-based reward}) can be computed by
\begin{equation}
r_t^i=\eta D_{KL}\left[q(\theta;\psi_{t+1})\parallel q(\theta;\psi_t)\right].
\end{equation}

Remarkably, VIME outperforms heuristic exploration models in different continuous control tasks. Meanwhile, the prioritization of curiosity-driven exploration can be controlled by scaling the intrinsic rewards. Nevertheless, a notable shortcoming is the instability of model performance, with a 25\% chance of failure in the half-cheetah task due to random seeds \cite{houthooft2016vime} and a 46\% of failure in an unseen maze by poor generalization \cite{DBLP:conf/iclr/GoyalISALBBL19}. In addition, VIME only focuses on robotic locomotion problems and agent can only learn one single policy rather than multiple skills across various domains. A possible way to stabilize the training process and avoid randomness might be to ensemble multiple dynamics models and therefore obtain the ``uncertainty" of uncertainty, with a similar idea to the work  \cite{pathak2019self}. However, an obvious trade-off would be the introduced training complexity caused by the ensemble. 

By extending the idea of uncertainty reduction in high-dimensional and complex environments, \citeauthor{berseth2020smirl}~\cite{berseth2020smirl} utilized Variational Autoencoders (VAE) to learn a non-linear state representation and proposed SMiRL that intrinsically rewards agents to remain in safe and stable states. In other words, rewards are given for minimizing the state entropy. In particular, the entropy of the state marginal distribution $H(d^\pi (s))$ is bounded and can be estimated by the distribution of visited states $p_\theta(s)$ parameterized by $\theta$.
\begin{equation}
\label{smirl}
\sum_t H(d^\pi (s_t))= - \sum_t  \mathbb{E}_{s\sim d^\pi (s_t)} \left[ \log d^\pi (s_t)\right]
\leq -\sum_t \mathbb{E}_{s\sim p_\theta (s_t)} \left[ \log p_\theta (s_t) \right].
\end{equation}
Following the objective of reducing actions to lower the state entropy in stochastic environments, minimizing the right hand side of Eq. (\ref{smirl}) is equivalent to incorporating an intrinsic reward
$r^i_t=\log p_{\theta_{t-1}} (s_t)$. As a result, the agent is capable of learning emergent behaviors such as keeping the balance during walking, directly from raw pixels (aided by VAE) and even without $r^e_t$ given. Based on SMiRL, \citeauthor{chen2020reinforcement}~\cite{chen2020reinforcement} focused on minimizing the future entropy to obtain more robustness and better generalization capability in procedurally generated environments. However, this type of CDL method is only applicable to certain RL tasks that only prefer avoiding catastrophic outcomes, which cannot be generalized to other tasks with thorough exploration and complex controls. Moreover, the choice of the density model may affect the agent's performance to some extent and the trained agent would rather stay in a highly predictable (low-entropy) part of the state space when the environment is only partially observed or highly stochastic.

To balance the exploration and exploitation trade-off in a partially observed task, \citeauthor{fickinger2021explore}~\cite{fickinger2021explore} recently constructed two adversarial policies that compete with each other: an Explore policy with $r^i_t=-\log p_{\theta_t} (o_{t+1})$ and a Control policy with $r^i_t=\log p_{\theta_t} (o_{t+1})$. By leveraging the adversarial technique, agents can efficiently master a complex control task by self-proposing a curriculum that becomes increasingly challenging. Yet, it remains challenging to quickly adapt to various downstream tasks with limited environment interactions.

Apart from model-based RL discussed above, uncertainty-based curiosity can be incorporated into model-free RL algorithms as well. Soft Actor-Critic (SAC) \cite{haarnoja2018soft} is an off-policy model-free RL algorithm that learns a stochastic policy $\pi_\psi$ (actor) with state-action value functions $Q(s,a)$ (critics), and a temperature coefficient $\alpha$ by encouraging exploration through a $\gamma$-discounted maximum-policy-entropy term. The key idea of entropy-maximization is to freely explore all possible actions and trajectories that are uncertain. Prior works have shown that maximizing entropy can help agents to improve exploration with diverse behaviors \cite{haarnoja2017reinforcement,o2018uncertainty} and learn a robust policy against model errors \cite{ziebart2010modeling,nachum2017trust}. The conventional Q-learning based RL algorithm defines the critics $Q$ based on the Bellman Equation
\begin{equation}
Q(s_t,a_t)=r^e(s_t,a_t)+\gamma \mathbb{E}_{s_{t+1}\sim p} \left[ V(s_{t+1})\right] \text{ with } V(s_{t+1})=\mathbb{E}_{a_{t+1} \sim \pi} \left[Q(s_{t+1},a_{t+1})\right].
\end{equation} 
Based on that, SAC motivates the agents for exploring trajectories with high uncertainty under the current policy by updating $V(s_{t+1})=\mathbb{E}_{a_{t+1} \sim \pi} \left[Q(s_{t+1},a_{t+1})-\alpha \log \pi (a_{t+1}|s_{t+1})\right]$ for the soft-Q function, which is equivalent to defining 
\begin{equation}
r^i=\alpha H(\pi(\cdot|s_{t+1}))=-\alpha \mathbb{E}_{a_{t+1} \sim \pi} \left[\log \pi (a_{t+1}|s_{t+1})\right].
\end{equation} 
During policy improvement, the KL divergence between the exponential of the soft-Q function and the policy is minimized to ensure $Q^{\pi_{new}}(s,a) \geq Q^{\pi_{old}}(s,a), \forall s, a$. Compared to the other RL models that learn a deterministic policy, SAC learns an optimal stochastic policy by maximizing the action entropy. More near-optimal policies and behaviors can be learned and utilized to more complex and specific downstream tasks. Since agents need to explore the environments under possible scenarios, the agents can therefore perform more robustly with a desirable generalization capability. However, SAC cannot learn directly from raw pixels and sample efficiency has become another challenge. Therefore, more recent works have been focusing on resolving this issue with SAC as a base learner \cite{laskin2020reinforcement,laskin2020curl,yarats2020image,yarats2021improving,nguyen2021sample}. To further balance the trade-off between exploration and exploitation, \citeauthor{lin2020cat}~\cite{lin2020cat} proposed an instance-level entropy temperature for SAC, which adaptively allows more exploration by a large entropy in unfamiliar states and more exploitation by a small entropy in familiar states.

So far, we have assumed that there is no prior information or experiences available, so agents need to learn and explore from scratches. In contact, it is easier to obtain examples of successful outcome states in some practical tasks such as navigation and robotic manipulation, where the uncertainty-based curiosity can shape a sophisticated intrinsic reward for more effective guidance towards the target goal. Under this problem setting, MURAL \cite{li2021mural} constructed a Bayesian classifier $p_\phi$ to measure the uncertainty of whether a particular state leads to the successful outcome based on the conditional normalized
maximum likelihood distribution \cite{rissanen2007conditional,zhou2021amortized}. The measured uncertainty is directly used as $r^i$ to replace $r^e$ and incentivize exploration with pure curiosity
\begin{equation}
r^i(s)= \frac{p_{\phi_1} (\text{success} | s)}{p_{\phi_1} (\text{success} | s)+p_{\phi_0} (\text{unsuccess} | s)},
\end{equation}
where $\phi_0$ and $\phi_1$ are the classifier parameters for adding $s$ into the data sets of successful outcomes and unsuccessful outcomes, respectively. A notable limitation is the required computational cost for extensively updating the classifiers. Thus, the meta-learning technique \cite{finn2017model} is employed to aid it. As a result, the proposed uncertainty-based curiosity is accurate and well-shaped, compared to other uncertainty measurements such as MLE and regularized MLE. With only $r^i$, agents can master difficult navigation and robotic manipulation tasks and obtain state-of-the-art performance. However, MURAL is only applicable to the tasks with successful outcome examples given and it remains to extend MURAL to MDP with only visual observations or partially observed problem settings.

\subsection{Novelty-Based Curiosity}
As discussed in Section \ref{collative variable}, novelty acts as a collative variable for arousing curiosity, which could be inversely measured by the frequency (i.e. density) of encountering the same (similar) stimulus. Motivated by novelty-based curiosity, agents are expected to explore \textit{novel} states and learn new behaviors (i.e. discouraged from revisiting the same states). Counting state, action, or state-action visitations is a direct way for measuring \textit{novelty} of the state, where a high count means a frequent visitation of the same state, action, or state-action and therefore low novelty and less curiosity.  

Over the past few years, it has been theoretically-justified that count(novelty)-based curiosity could enhance the performance in RL agents. For instance, the Upper-Confidence-Bound (UCB) strategy \cite{auer2002finite} intrinsically incentivizes the agents to choose the under-explored action under the bandit setting, 
\begin{equation}
a_t= \arg \max_a \left[Q_t(a) + \alpha \sqrt{\frac{2\ln t}{N_t(a)}}\right],
\end{equation}
where $Q_t(a)$ is the estimated value of action $a$ controlling the exploitation, $N_t(a)$ represents counting of $a$ actions prior to time step $t$ and $\alpha$ is the temperature coefficient. The second term effectively encourages exploration in a curious manner. If an action has not been selected quite often i.e. $N_t(a)$ is small, the second term will become large and agents will be more likely to try this action. With the increment of $N_t(a)$ and time progress, the action becomes less novel due to frequent visitation and exploration. Therefore, agents shall concentrate less on the familiar actions and leverage more exploitation due to being less curious. By extending the counting of actions to state-action pairs, the Model Based Interval Estimation–Exploration Bonus (MBIE-EB) \cite{strehl2008analysis} utilized $r_t^i(a_t)=\alpha N(s,a)^{-\frac{1}{2}}$ to encourage exploring less visited state $s$ and action $a$ pair with a state-action count $N(s,a)$. Similarly, In the Bayesian Exploration Bonus (BEB) algorithm \cite{kolter2009near}, actions are determined based on an additional bonus (intrinsic reward) of $r_t^i(a_t)=\frac{\alpha}{1+N(s,a)}$ for infrequent observed state-action pairs. However, all these methods require the state (or state-action) space to be low-dimensional and countable to ensure most states will not only occur once, which has become problematic for continuous high-dimensional control tasks.

A main challenge is that the number of occurrences $N_t(s)$ of a yet-unseen state $s$ in $\{s_1,s_2,\cdots,s_t\}$ is generally zero for continuous controls.  To address the above issue, \citeauthor{bellemare2016unifying}~\cite{bellemare2016unifying} proposed a count-based exploration method where the intrinsic reward is determined by the pseudo-count function $\hat{N}_t(s)$ rather than $N_t(s)$. In particular, the pseudo-count $\hat{N}_t(s)$ can be explicitly obtained through a density model $p$ over the state space $\mathcal{S}$, which can be expressed as
\begin{equation}
\hat{N}_t(s)=\frac{p_t(s)(1-p'_t(s))}{p'_t(s)-p_t(s)},
\end{equation}
where $p_t(s)$ and $p_t'(s)$ (recoding probability) are respectively defined as
\begin{align}
\label{rho}
&p_t(s)=P(S_{t+1}=s | S_1=s_1,\cdots,S_t=s_t) \nonumber
\\&p_t'(s)=P(S_{t+2}=s | S_1=s_1,\cdots,S_t=s_t, S_{t+1}=s) 
\end{align}

Furthermore, the authors have shown the relations among information gain $IG$, prediction gain $PG$ and pseudo-count $\hat{N}_t(s)$: $IG_t(s)\leq PG_t(s)\leq \hat{N}_t(s)^{-1}$ and $PG_n(s)\leq \hat{N}_n(s)^{-\frac{1}{2}}$. We refer readers to the rigorous proof provided in the paper \cite{bellemare2016unifying}, which links novelty-based curiosity and uncertainty-based curiosity theoretically. Most importantly, it has implied that using an intrinsic reward proportional to $\hat{N}_n(s)^{-\frac{1}{2}}$ can lead to a behaviour at least as exploratory as one derived from information gain in VIME \cite{houthooft2016vime}. Therefore, a particular $r_n^i=\alpha (\hat{N}_n(s)+0.01)^{-\frac{1}{2}}$ was employed in the experiments. The experimental results showed that count-based intrinsic reward could obtain significantly improved exploration in various hard games such as Atari \cite{mnih2013playing}. However, it did not consider the action space to be continuous and the states are assumed to be acquired directly. In more practical applications, agents should learn a meaningful state representation directly from raw pixels without causing the density model to become prohibitively expensive. 

For the density model $p$, \citeauthor{ostrovski2017count}~\cite{ostrovski2017count} further argued that several critical assumptions for training neuronal density model must be satisfied during implementation:
\begin{itemize}
	\item $p$ should be learning-positive, i.e. $p'_t(s)\geq p_t(s)$.
	\item $p$ should be trained completely online, using each state $s$ exactly once given the sequential order.
	\item The prediction gain should decay at a rate of $t^{-1}$ to ensure that $\hat{N}$ grows approximately linearly with $N$.
	\item Training of $p$ should be done by minibatches, drawn randomly from a diverse dataset to avoid catastrophic forgetting in case of a drifting data distribution.
	\item The choice of density model must be computationally efficient enough to accommodate prediction gain update in the RL framework.
\end{itemize}
To avoid substantially increasing the model complexity, they employed PixelCNN for the density model which has obtained greater accuracy, faster learning speed, higher stability, and wider compatibility with various RL algorithms. Meanwhile, the mixed Monte Carlo update is found to be a powerful facilitator of effective exploration. However, the sample efficiency still needs to be improved when only limited samples are collected and it cannot perform well for continuous control tasks.

To further reduce the model complexity for measuring the novelty(count)-based, \citeauthor{tang2017exploration}~\cite{tang2017exploration}proposed an encoded hash function $h(s)$ that can map the high-dimensional continuous state space into a low-dimensional feature space. Thus, $s\in\mathcal{S}$ can be discretized by a hash-based pseude-count, which is used to formulate an intrinsic reward
\begin{equation}
r_t^i={\alpha}{N(h(s))}^{-\frac{1}{2}}.
\end{equation}
While the simple hash functions have obtained comparable results in many challenging tasks, the domain-dependent learned hash code via locality-sensitive hashing could make further improvement. Overall, it is a simple generalized method to reach near state-of-the-art performances across different domains. However, the trade-off is that training autoencoder-based hash code will require a large number of parameters to be updated, so the learning process will be slowed down. In addition, the agents did not perform well at the early stage of training, indicating a poor sample efficiency.

To avoid using generative models for density estimation and therefore reduce the computational cost, a scalable exploration algorithm named EX$^2$ is proposed by training a discriminative exemplar model $D_s(s)$ for the newly visited state against all past seen states~\cite{fu2017ex2}. If a newly visited state can be easily distinguished from the past states, it therefore will be considered relatively novel. Therefore, the exemplar model can be utilized to implicitly estimate the state density and approximate the pseudo-count novelty for shaping an intrinsic reward
\begin{equation}
r^i(s,a)=-\alpha \log D_s(s).
\end{equation}
Remarkably, the density estimation is performed by $D_s(s)$, which avoids reconstructing samples to train and has outperformed the prior generative methods on domains with more complex observation functions. As this method incorporates the adversarial mechanism, the instability issue may occur during training.

Measuring ``novelty” by count-based methods in RL requires constructing or approximating a statistical model to describe the environmental state distribution, which is still an open problem for dealing with complex and high-dimensional observations. Apart from the count-based approach, novelty can also be measured implicitly through reachability. Given an encoded state $e$ from a visual observation $o$, the estimated environment steps to take from the experienced states stored in a memory buffer $M$ to the given $e$ can be used to describe the degree of novelty for $o$ \cite{savinov2018episodic}. Thus, those observations which require substantial efforts (high novelty) to reach will be given high intrinsic rewards for exploration
\begin{equation}
r^i=\alpha (\beta + C(M,e)),
\end{equation}
where $\alpha$ and $\beta$ are hyper-parameters, and $C(M,e) \in [0,1]$ represents the reachability of $e$. This reachability-based novelty effectively encourages agents to outperform other models in visually rich 3D environments by avoiding unproductive noises (``couch potato" issue). In more practical scenarios where only partial observations or limited samples are available, it is unclear whether this model can still produce desirable performances.

So far, the novelty is explicitly or implicitly measured on whole states or observations. In practice, not all information contained in the observations is necessary for evaluating the task-dependent novelty, so the noisy task-irrelevant information should be excluded. For this purpose, \citeauthor{DBLP:conf/icml/KimNKKK19}~\cite{DBLP:conf/icml/KimNKKK19} utilized the variational information bottleneck framework \cite{DBLP:conf/iclr/AlemiFD017} to learn the compressive and informative representation $Z$ for quantifying task-relevant novelty. Given the task information $Y$ and observation $X$, $Z$ is learned by minimizing $-I(Z;Y)+\beta I(X;Z)$, where $I(\cdot)$ is the mutual information and $\beta$ is an importance weight. As $I(X;Z)$ is intractable, the variational inference by $q(Z)$ is employed and the intrinsic reward is represented with a KL-divergence term
\begin{equation}
r^i(x_n)=D_{KL}\left[p(Z|x_n)||q(Z)\right].
\end{equation}
The proposed $r^i$ desirably measures the task-related state novelty and the theoretical proof is provided in the work \cite{DBLP:journals/corr/abs-1807-00906}. By rewarding agents to explore novel states related to the given tasks, agents can obtain surpassing performances without distraction than previous works \cite{tang2017exploration,burda2018exploration}. However, the importance weight $\beta$ that balances between compression and preservation of information has to be carefully designed for different tasks, where an adaptive generalization is more desirable.

In addition to encouraging exploration, curiosity can also enhance sample efficiency in the sense that novel experiences should be replayed more frequently so that agents can learn an optimal policy more efficiently. The conventional off-policy RL algorithms often construct a replay buffer $\mathcal{B}$ to store a large amount of transition tuples $\tau_t=(s_t, a_t, r_t, s_{t+1})$. During the model training, a batch of $\tau_t$ are sampled uniformly from $\mathcal{B}$ and feed into the RL learner. To direct agent's attention on under-explored (novel) experiences, \citeauthor{zhao2019curiosity}~\cite{zhao2019curiosity} proposed a curiosity-driven prioritization (CDP) framework that encourages to over-sample the trajectories with rare achieved goal states. The trajectory density $\rho(\mathcal{T})$ is measured by fitting a normalized Variational Gaussian Mixture Model (V-GMM) using the replay buffer every epoch. During experience replay, the probability of a trajectory $\mathcal{T}$ to be sampled with the prioritization can be expressed as
\begin{equation}
p(\mathcal{T})=\frac{\text{rank}(\bar{\rho}(\mathcal{T}))}{\sum_i \text{rank}(\bar{\rho}(\mathcal{T}_i))}, \text{ with } \bar{\rho} \propto  1- \rho.
\end{equation}
Compared to other experience replay prioritization methods based on temporal difference (TD) error \cite{schaul2016prioritized}, the CDP is more efficient by avoiding extensive updates of TD errors. By training in different robot manipulation tasks, CDP can improve both sample efficiency and model performance. However, it requires the extrinsic reward $r^e$ to be well defined and continuously given. In the absence of extrinsic reward, CDP might not be able to learn an optimal policy. Meanwhile, CDP only considers the MDP process with states rather than visual observations, and fitting a V-GMM on raw pixels directly is rather challenging and time-consuming.

\subsection{Incongruity-Based Curiosity}
In some environments, salient events such as light switching on/off and music turning on/off may occur unexpectedly, which indicates the informative knowledge about the actions taken or states visited that has yet been learned. Following the definition in Section \ref{collative variable}, those \textit{unexpected} salient events may induce incongruity-based curiosity and it was proposed that these events can be leveraged to formulate an incongruity-based reward to learn the novel and meaningful information \cite{barto2004intrinsically}. By constructing a prediction model, the unexpected salient events can be identified and the degree of un-expectation (\textit{incongruity}) can be utilized as a direct quantification of the intrinsic reward, which encourages agents to actively update their internal beliefs.  

To extend the motivation of incongruity-based curiosity, \citeauthor{chentanez2004intrinsically}~\cite{chentanez2004intrinsically} proposed a computational RL algorithm utilizing the options framework $\mathcal{O}$, where the probability of not expecting the salient event is formulated as the intrinsic reward
\begin{equation}
r_{t+1}^i = \left\{
\begin{array}{ll}
\alpha [1-P^{O_e}(s_{t+1}\mid s_t)] & \quad \text{if } s_{t+1}\text{ is salient} \\
0& \quad \text{otherwise}
\end{array}
\right..
\end{equation}
where $\alpha$ is a scaling factor, and $P^{O_e}(s_{t+1}\mid s_t)$ is the probability of terminating at $s_{t+1}$ under the current option $O_e$ given $s_t$. 
When the agent encounters an unexpected salient event a few times, the intrinsic reward diminishes with an improved option policy and policy model. Subsequently, the agent becomes ``bored” with the associated salient event and moves on. Therefore, the agent is able to learn efficiently and perform well with the proposed incongruity-based intrinsic rewards. Nevertheless, one crucial assumption is that the salient event must be observable and deterministic, so it might be difficult to identify salient events in complex environments or apply this method to more general tasks without salient events.

\subsection{Surprisingness-and-Incongruity Based Curiosity}
As introduced in Section \ref{collative variable}, psychological curiosity can be evoked through surprisingness and incongruity. In RL, these two collative variables are often integrated into a single concept of deviation from the agent's internal belief such as \textit{prediction errors}. The majority of existing works have designed various intrinsic rewards based on prediction errors to motivate agents to explore the under-explored states that contract to their internal beliefs the most and update their knowledge accordingly. On the other hand, experience replay prioritization based on temporal difference (TD) errors enables agents to concentrate more on the under-explored trajectories and update the internal belief repeatedly. As a result, the prediction errors and TD errors can be reduced continuously and the agent is able to forecast the consequence of its own actions more accurately, which is a desirable feature in many applications. 

To avoid the construction of the environmental dynamic model that predicts the next state $s_{t+1}$ given the action $a_t$ executed in the current state $s_t$, \citeauthor{huang2002novelty}~\cite{huang2002novelty} directly utilized the sensation difference between the robot's prediction and what the robot actually sensed as the intrinsic reward to enhance the learning capacity,
\begin{equation}
r_t^i=\sqrt{\frac{1}{m}\sum_{j=1}^{m}\frac{(\hat{s}_{t+1}^{(j)}-s_{t+1}^{(j)})^2}{\sigma^{(j)}_{t}}},
\end{equation}
where $\hat{s}_{t+1}^{(j)}$ represents the primed sensation for the next scene in the $j$-th dimension, $m$ represents the dimension of sensory input and $\sigma^{(j)}_{t}$ represents the expected deviation for the $j$-th dimension, which could be computed by the time-discounted average of the squared difference $(\hat{s}^{(j)}-s^{(j)})^2$.  The proposed intrinsic rewards enhanced the learning capacity for robots; however, a notable shortcoming is the prohibitively inefficient performance when dealing with the high-dimensional space or stochastic dynamics. 


To efficiently compute the prediction error in high-dimensional space containing irrelevant information and obtain a better generalization capacity in unseen scenarios even when $r_t^e$ is completely absent, a curiosity-driven exploration process governed by the self-supervised prediction is proposed \cite{pathak2017curiosity}. Its intuition is to intrinsically incentivize the agents to explore in a curious manner when the predicted next states differ from the actual ones. In particular, intrinsic rewards are generated through the Intrinsic Curiosity Module (ICM), consisting of a feature model $\phi$ to extract meaningful state presentations from high-dimensional continuous state spaces (such as raw pixels), an inverse model $\hat{a_t}$ to approximate the action taken and optimize $\phi$, and a forward model $\hat{\phi}$ to obtain the expected feature in the next state. 

By jointly optimizing all components of ICM, the extracted features from high-dimensional states are meaningful and robust to the uncontrollable noise such as pure random movements in the background image. Therefore, the difference between actual and predicted features of the next state is reliable and can be utilized as an intrinsic reward
\begin{align}
r^i_t=\frac{\alpha}{2} \|\phi({s_{t+1}})-\hat{\phi}_{s_{t+1}}\|^2_2.
\end{align}
By conducting experiments on Super Mario Bros and VizDoom, the curiosity-driven exploration has improved sample efficiency of learning directly from raw pixels and the agent is capable of learning an optimal policy even without the extrinsic rewards given. The learned exploratory behaviour by the curious agent is not only specific to training space, but also can be generalized to perform in unseen scenarios. As ICM is capable of learning directly from raw pixels, \citeauthor{yu2020intrinsic}~\cite{yu2020intrinsic} proposed to integrate ICM with generative models for imitation learning in a high-dimensional environment, and the agent can outperform experts by generating more state-action pairs and more accurate rewards. Nevertheless, one problem of ICM is that the intrinsic reward may vanish during training, which cannot encourage the agent to further explore the environment. Meanwhile, the model performance can be unstable by training with different random seeds.  

To further stabilize the model performance by enhancing the reliability of prediction errors, \citeauthor{pathak2019self}~\cite{pathak2019self} utilized multiple dynamics models $\hat{\phi}$ and employed the variance of model predictions $\hat{\phi}_{s_{t+1}}(s_t, a_t)$ as the intrinsic reward
\begin{equation}
\label{variance}
r^i_t = \text{var}(\hat{\phi}_{s_{t+1}}(s_t, a_t))=\mathbb{E}\left[ (\hat{\phi}_{s_{t+1}}(s_t, a_t)-\phi({s_{t+1}}) )^2\right].
\end{equation}
It can also be extended to the multi-agent exploration problem as a joint curiosity reward shared by all agents \cite{yang2021ciexplore}. Intuitively, consistent predictions of the next states indicate low curiosity as the state space has been well explored by the agent with low uncertainty. In contrast, agents become curious when a disagreement arises on the next state prediction among the ensemble of dynamics models. In this case, agents will be encouraged to explore the novel areas that have caused high prediction errors. Compared to the single forward dynamics model, using the variance of prediction errors as an intrinsic reward can enable agents to explore in stochastic environments. However, the computational complexity increases with the ensemble amount of dynamics models and it requires exploring extensively across all state space to learn an optimal policy for a single downstream task.

By integrating the ensemble idea for exploration~\cite{pathak2019self} and planning, \citeauthor{sekar2020planning}~\cite{sekar2020planning} proposed a Plan2Explore model to learn a world model about its environments and subsequently solve zero-shot and few-shot downstream tasks. The exploration is carried out by utilizing Eq. (\ref{variance}) as an intrinsic reward in a self-supervised manner. Afterward, the agent receives reward functions and is able to quickly adapt to various tasks such as walking, running and balancing in the DeepMind Control Suite \cite{tassa2018deepmind}. Theoretically, this work has shown maximizing the variance of ensemble means is approximately maximizing the information gain as introduced in Section \ref{Uncertainty-Based Reward}. To achieve desirable performance by quick adaption, the world model must be trained with sufficient exploration, where the sample efficiency can be further improved.

However, the choice of feature embedding in the ICM plays an important role in the CDL performance. Given any high-dimensional state, the feature model should capture two types of features: (1) features that can be affected by the agent and (2) features that cannot be controlled by the agent but can affect the agent (e.g. the features of a vehicle driven by another agent). Meanwhile, the feature model will filter out (exclude) the features that cannot be controlled by the agent or cannot affect the agent (e.g. moving clouds in the background). 
To investigate the effects of different feature embedding models, \citeauthor{burda2018large}~\cite{burda2018large} evaluated ICM \cite{pathak2017curiosity} in 54 environments and summarized three critical criteria (compact, sufficient and stable) for constructing and evaluating the desired embedding feature model. 
\begin{itemize}
	\item Compact: The learned features should have low dimensionality with irrelevant elements of the observation space being filtered out.
	\item Sufficient: The learned features should preserve all essential information about the state. Otherwise, no intrinsic reward can be effectively identified when the agent explores some relevant elements in the environment.
	\item Stable: The learned features should minimize the changes with the same observations during the learning process, thus providing stationary rewards. Otherwise, the proposed curiosity can become misleading and inconsistent.
\end{itemize}
In addition, four types of common feature embedding methods including Pixels, Random Features (RF) (similar to random network distillation \cite{burda2018exploration}), VAE, and Inverse Dynamics Features (IDF) are employed to compare the CDL performances. Table \ref{table:compare} indicates their different characteristics and performance, which provides a justification for proper method selection in the future. While IDF is more capable of generalizing to unseen observations, it still suffers from the ``couch potato" issue (``noisy TV" problem) with  significant drops in agent's learning performances.

\begin{table}[]
	\centering
	\scalebox{0.9}{
		\begin{tabular}{p{6cm}|cccc}
			\toprule
			& \textbf{Pixels}	& \textbf{RF}  & \textbf{VAE}	& \textbf{IDF}   \\ \midrule		
			Compact
			&No
			&Maybe
			&Yes
			&Yes\\
			Sufficient
			&Yes
			&Maybe
			&Yes
			&Maybe\\
			Stable
			&Yes
			&Yes
			&No
			&No\\
			\small Performance on simple env.
			&Bad
			&Good
			&Good
			&Good\\
			\small Performance on complex env.
			&Bad
			&Average
			&Good
			&Good\\
			\bottomrule
	\end{tabular}}
	\caption{A comparison of different feature embedding methods: Pixels cannot encode a compact state representation, resulting in worst performances. RF cannot fully guarantee to be compact and sufficient, thereby its performance can vary in challenging environments. Although VAE and IDF are not stable at the early stage of learning, they can be jointly trained with RL and gradually become stabilized. Therefore, both VAE and IDF perform relatively well across different tasks.
	}
	\label{table:compare}
\end{table}

To improve sample efficiency by obtaining a meaningful state representation, contrastive learning is utilized with a forward dynamics model in CCFDM to carry out self-supervised representation learning \cite{nguyen2021sample}. In particular, the prediction error of whether the augmented current observations $q$ can infer the state information of the augmented next observations $k$ accurately is employed as an intrinsic reward
\begin{equation}
r_t^i=\alpha \exp (-\eta t) \text{ dissim}(q,k)\frac{r^e_{\max}}{r^i_{\max}},
\end{equation}
where $\alpha$ is a temperature coefficient, $\eta$ is the decay weight, $r^e_{\max}$ and $r^i_{\max}$ are respectively the maximum extrinsic rewards and intrinsic rewards.
In this formulation, $r_t^i$ is standardized among different tasks and gradually decreased. The agent is encouraged to explore the states where the state representation is inaccurate. As a result, the sample efficiency for continuous control tasks has been enhanced in a self-supervised way but the model performance is rather unstable with a high variance of performance scores.

Besides intrinsic rewards, the prediction error can also be used in the prioritized experience replay through temporal difference (TD) errors. For the transition tuples $\tau_t=(s_t, a_t, r_t, s_{t+1})$ stored in the replay buffer, they are not equally important during learning. In other words, there must be some tuples that can help agents to converge faster. In particular, agents should be curious about the tuples with high TD errors $\delta=\left[r_t^e+\gamma Q(s_{t+1},\pi(s_{t+1}))\right]-Q(s_t,a_t)$ as they may contain novel knowledge that has yet been mastered in Q-learning. In the prioritized experience replay (PER) \cite{schaul2015prioritized}, the probability of tuple $i$ being sampled for replay $P_i$ and its corresponding sample-importance weight $w_i$ can be formulated as follows,
\begin{equation}
P_i = \frac{p_i^\alpha}{\sum p_i^\alpha} \text{ with } p_i = \frac{1}{\text{rank}(\delta_i)}, ~w_i=\frac{(N\cdot P_i)^{-\beta}}{\max w_i},
\end{equation}
where $\alpha$ is a constant, $N$ is the total number of tuples in the replay buffer and $\beta$ is a hyper-parameter controlling the degree of offsetting. Therefore, all tuples have different probabilities to be sampled for improving the training speed; meanwhile, $w_i$ corrects the introduced bias due to the change in distribution.
Rainbow DQN \cite{hessel2018rainbow} further integrated this idea with other merits of DQN variants. The agents have obtained outstanding performance that is superior than human players in Atari environments \cite{mnih2013playing}. However, TD-error-based prioritization does not explicitly improve representation learning and can become misleading when the learned state is inaccurate. During training, it requires extensive evaluations of TD errors and importance-sampling weights, increasing the computational cost to some extent. 

Recently, variants of TD-error-based prioritization methods have been proposed to further yield training efficiency. For example, \citeauthor{kapturowski2018recurrent}~\cite{kapturowski2018recurrent} proposed to utilize a mixture of max and mean absolute n-step TD-errors over the sequence, which can effectively avoid large errors and exploit long temporal information. Similarly, \citeauthor{brittain2019prioritized}~\cite{brittain2019prioritized} extended PER to increase the priority of the previous transition tuples that lead to the ones with high TD errors, theoretically converging faster than PER. However, these methods are at the cost of increasing sample complexity and sample efficiency remains challenging.

\subsection{Complexity-Based Curiosity}
Given procedurally-generated environments with a parameterized family of tasks, it is infeasible to perform various skills directly with sparse extrinsic rewards. By identifying the 
\textit{complexity} of goals and seeking the optimal level of complex tasks, agents can be continually motivated to learn with high curiosity, which is consistent with the arousal mechanism as introduced in Section \ref{framework}. In AMIGO, \citeauthor{campero2020learning}~\cite{campero2020learning} proposed to self-generate complex goals (coordinates) through a ``teacher" module while a goal-conditioned “student” policy learns to complete these tasks in a curious manner. During the constructively adversarial training, the student is rewarded by $r_{student}^i=1$ when reaching the teacher-proposed goal coordinates. Meanwhile, the teacher is rewarded by generating complex goals that are not too easy or too challenging for the student to complete $r_{teacher}^i= \left\{
\begin{array}{ll}
a & \quad t^+ \geq t^* \\
-b & \quad  t^+ < t^*
\end{array}
\right.$, where $a,b \in \mathbb{R}^+$, $t^+$ measures the steps that were taken by the student to reach the goal, and $t^*$ is the threshold that increases as the student improves. In this way, agents can curiously carry out meaningful exploration with a natural curriculum and learn a diverse range of skills in procedurally-generated environments. However, the generated goal is limited to coordinates and more abstract forms of goals in more rich domains are desirable.

\subsection{Change-Based Curiosity}
As introduced in Section \ref{collative variable}, curiosity can be aroused by the change of stimulus. In RL, this collative variable refers to the substantial state \textit{change} after agents take a particular action. Intuitively, agents become curious when their actions lead to significant changes in the environment, thereby learning impactful skills to control the environment. By utilizing ICM \cite{pathak2017curiosity} that learns a meaningful state representation, \citeauthor{DBLP:conf/iclr/RaileanuR20}~\cite{DBLP:conf/iclr/RaileanuR20} proposed RIDE with a change-based intrinsic reward to attract agents to explore the unfamiliar states where they can greatly influence through substantial changes in the learned state representation. The state representation $\phi$ is trained by ICM to extract features controllable by the agent and the non-controllable features that can affect the agent. Thus, the intrinsic reward can be formulated by
\begin{equation}
r_t^i=\frac{||\phi(s_{t+1})-\phi(s_{t})||_2}{\sqrt{N_{ep}(s_{t+1})}},
\end{equation}
where $N_{ep}(s_{t+1})$ is the visited times for countable state or episodic pseudo-counts for high-dimensional states. The numerator encourages agents to take the actions with substantial influences while the denominator prevents agents from repeatedly exploring the same states. Meanwhile, this intrinsic reward does not diminish to zero, guiding agents to learn even after training for a long time. However, RIDE is only applicable to the tasks where agents need to affect the environments. For the tasks requiring agents to keep stable (balancing, walking, and running), actively seeking significant change may not be suitable.   

\subsection{Discussion}
To summarize, curiosity-driven learning has been widely applied in RL and we focus on the important role of  curiosity with a majority in intrinsic rewards and few works in experience replay prioritization. A taxonomy of the current literature on curiosity-driven RL is shown in Table~\ref{fig:taxonomy_RL}. Given different challenging problems such as sparse extrinsic rewards, learning directly from raw pixels, and low sample efficiency, curiosity can incentivize agents to explore novel states and reduce the uncertain/prediction error towards unseen states, thereby quickly learning optimal policies or mastering necessary skills under a diverse range of task environments. Nevertheless, there is abundant potential to further improve the curiosity-driven RL, which can be summarized as follows. 
\begin{itemize}
	\item The existing forms of curiosity in RL are almost inspired by only one type of collative variables with a linear relation. However, it is desirable to integrate more collative variables that can comprehensively quantify the curiosity with a non-linear relation from the different perspectives. 
	\item Most CDL methods have enabled agents to learn a single policy or skill with enhanced learning performance. To enhance the generalization capability, it is possible to extend the learned primitive/low-level policy in a more complex and hierarchical architecture such as a meta-learning system that can quickly adapt to different downstream tasks. 
	\item For high-dimensional continuous spaces, learning a meaningful state representation with further improved sample efficiency remains a challenging task for efficient RL. To remedy this, utilizing a stable CDL framework may help to improve representation learning, which can be model-agnostic in the sense that it is compatible with most RL base learners to learn from high-dimensional continuous spaces. 
	\item Most of the proposed curiosity tend to diminish nearly to 0 at the later stage of training. Although curiosity level generally reduces w.r.t. the learning progress, it should be able to increase whenever a new external goal or downstream task is provided. Therefore, it remains challenging to propose an adaptive curiosity that can quickly motivate agents to solve a new task and attain human-like learning capabilities.
	\item Naive CDL methods might be vulnerable to the ``couch potatoes" issue (noisy TV problem), where curious agents can be easily distracted by task-irreverent novelty, unpredictable noise, or inherently high-entropy. In general, highly stochastic dynamics can drastically affect learning performances, which should be addressed by a more nuanced CDL to differentiate stochastic dynamics and uncertain dynamics. 
	\item According to the arousal mechanism framework in Section \ref{framework}, the majority of curiosity in this section are expected to be maximized for more effective and diverse explorations. However, this objective only corresponds to the reward system in Figure \ref{InvertedU} and only a few works \cite{fickinger2021explore} have considered the anxiety aversion system that the intensity beyond agent's manageable knowledge gap can negatively affect learning. Thus, it is desirable to investigate the effect of a more complete curiosity arousal mechanism by penalizing overly intensive stimulus and restricting the curiosity level within a certain range.
	\item When extrinsic rewards $r^e$ are available, most of the CDL methods simply sum up the formulated intrinsic rewards $r^i$ and extrinsic rewards $r^e$ through a fixed weighting coefficient. However, this weighting coefficient has to be hyper-tuned for different tasks, indicating a poor generalization capability. A direct intuition is to propose an adaptive scheduling of integrating extrinsic and intrinsic rewards, which may further regulate the learning process and yield an optimal balance between exploration and exploitation.
	\item It remains unclear whether the learned policy driven by various types of curiosity is sub-optimal or global optimal. Therefore, both empirical investigation and theoretical study are necessary in the future CDL research.
\end{itemize}

\begin{landscape}
	\begin{table}[]
		\tiny\sffamily\centering 
		\begin{center}
			{ \scalebox{0.95}{ \def\arraystretch{1.3}
						
				\begin{tabular}{|m{0.1\textwidth}|m{0.2\textwidth}|m{0.35\textwidth}|m{0.37\textwidth}|m{0.37\textwidth}|}
					\hline
					\textbf{Collative Variables}                           & \textbf{Models}                                                                                                                  & \textbf{Artificial   Curiosity}                                                                                                                                                                                                                                   & \textbf{Merits}                                                                                                                                                                                                                    & \textbf{Limitations}                                                                                                                             \\ \hline
					\multirow{5}{\linewidth}{Uncertainty}                    & VIME   \cite{houthooft2016vime}                                                                                   & $r_t^i=\eta   D_{KL}\left[q(\theta;\psi_{t+1})\parallel q(\theta;\psi_t)\right]$                                                                                                                                                                         & improve exploration   efficiency in continuous control tasks; avoid random walk behaviors                                                                                                                               & may perform unstably;   limited to robotic locomotion problems                                                                   \\ \cline{2-5} 
					& SMiRL \cite{berseth2020smirl}                                                                                     & $r^i_t=\log   p_{\theta_{t-1}} (s_t)$                                                                                                                                                                                                                    & master emergent   behaviors; learn from raw pixels; require no $r^e_t$                                                                                                                                                  & generalize poorly to   catastrophic-outcome-seeking tasks                                                                        \\ \cline{2-5} 
					& AS \cite{fickinger2021explore}                                                                                    & Explore policy:   $r^i_t=-\log p_{\theta_t} (o_{t+1})$; Control policy: $r^i_t=\log   p_{\theta_t} (o_{t+1})$                                                                                                                                            & solve partially   observed task; adversarially self-propose increasingly challenging goals                                                                                                                              & unable to quickly   adapt with limited environment interactions                                                                  \\ \cline{2-5} 
					& SAC \cite{haarnoja2018soft}                                                                                       & $r^i=\alpha   H(\pi(\cdot|s_{t+1}))=-\alpha \mathbb{E}_{a_{t+1} \sim \pi} \left[\log \pi   (a_{t+1}|s_{t+1})\right]$                                                                                                                                     & learn an optimal   stochastic policy for off-policy model-free RL; perform robustly with   desirable generalization                                                                                                     & perform poorly based   on raw pixels; learn with low sample efficiency                                                           \\ \cline{2-5} 
					& MURAL \cite{li2021mural}                                                                                          & $r^i(s)=   \frac{p_{\phi_1} (\text{success} | s)}{p_{\phi_1} (\text{success} |   s)+p_{\phi_0} (\text{unsuccess} | s)}$                                                                                                                                  & master difficult   navigation and robotic manipulation tasks; require no $r^e_t$                                                                                                                                        & require successful   outcome examples to be explicitly given; perform poorly from visual   observations                          \\ \hline
					\multirow{9}{\linewidth}{Novelty}                        & UCB   \cite{auer2002finite}                                                                                       & $\alpha   \sqrt{\frac{2\ln t}{N_t(a)}}$                                                                                                                                                                                                                  & \multirow{3}{\linewidth}{explicitly encourage explorations of less visited actions (or state-action)}                                                                                                                          & \multirow{3}{\linewidth}{limited to countable action   (state-action) space; cannot scale to continuous high-dimensional control   tasks} \\ \cline{2-3}
					& MBIE-EB \cite{strehl2008analysis}                                                                                 & $r_t^i(a_t)=\alpha   N(s,a)^{-\frac{1}{2}}$                                                                                                                                                                                                              &                                                                                                                                                                                                                         &                                                                                                                                  \\ \cline{2-3}
					& BEB \cite{kolter2009near}                                                                                         & $r_t^i(a_t)=\frac{\alpha}{1+N(s,a)}$                                                                                                                                                                                                                     &                                                                                                                                                                                                                         &                                                                                                                                  \\ \cline{2-5} 
					& Pseudo-Counts \cite{bellemare2016unifying}                                                                        & $r_n^i=\alpha   (\hat{N}_n(s)+0.01)^{-\frac{1}{2}}$                                                                                                                                                                                                      & encode continuous   state; theoretically prove its relation to VIME \cite{houthooft2016vime};                                                                                                                    & unable to learn from   raw pixels without causing the density model to become prohibitively   expensive                          \\ \cline{2-5} 
					& Hashing-Based Count \cite{tang2017exploration}                                                                    & $r_t^i={\alpha}{N(h(s))}^{-\frac{1}{2}}$                                                                                                                                                                                                                 & map the high-dimensional continuous state space into a low-dimensional feature space   via an encoded hash function                                                                                                   & train the autoencoder-based hash code slowly; generalize poorly at the early stage                                             \\ \cline{2-5} 
					& EX$^2$~\cite{fu2017ex2}                                                                                           & $r^i(s,a)=-\alpha   \log D_s(s)$                                                                                                                                                                                                                         & implicitly estimate   the state density via the adversarial mechanism                                                                                                                                                   & may perform unstably                                                                                                             \\ \cline{2-5} 
					& Reachability \cite{savinov2018episodic}                                                                           & $r^i=\alpha (\beta +   C(M,e))$                                                                                                                                                                                                                          & perform well in rich 3D environments; resolve the ``couch potato" (noisy TV) issue;                                                                                                                                  & may perform poorly in   partially observed tasks or few-shot settings                                                            \\ \cline{2-5} 
					& Curiosity-Bottleneck \cite{DBLP:conf/iclr/AlemiFD017}                                                             & $r^i(x_n)=D_{KL}\left[p(Z|x_n)||q(Z)\right]$                                                                                                                                                                                                             & rule out   task-irrelevant information during training                                                                                                                                                                  & unable to adapt the importance weight across different tasks                                                                  \\ \cline{2-5} 
					& CDP \cite{zhao2019curiosity}                                                                                      & $p(\mathcal{T})=\frac{\text{rank}(\bar{\rho}(\mathcal{T}))}{\sum_i   \text{rank}(\bar{\rho}(\mathcal{T}_i))}$                                                                                                                                            & avoid extensive updates of TD errors; improve sample efficiency through prioritized   experience replay                                                                                                               & requires $r^e$ to be well defined and continuously given; unable to learn efficiently from raw   pixels                        \\ \hline
					Incongruity                                     & Intrinsically   Motivated RL~\cite{chentanez2004intrinsically}                                                    & $r_{t+1}^i = \begin{cases}
					\alpha [1-P^{O_e}(s_{t+1}|s_t)] & \text{if salient}\\
					0 & \text{otherwise}
					\end{cases}$
					& leverage unexpected   salient events to improve the learning efficiency                                                                                                                                                 & require the salient   event to be observable and deterministic                                                                   \\ \hline
					\multirow{6}{\linewidth}{surprisingness and Incongruity} & SAIL~\cite{huang2002novelty}                                                                                      & $r_t^i=\sqrt{\frac{1}{m}\sum_{j=1}^{m}\frac{(\hat{s}_{t+1}^{(j)}-s_{t+1}^{(j)})^2}{\sigma^{(j)}_{t}}}$                                                                                                                                                   & avoid the   construction of environmental dynamic model; improve the learning capacity   for robots                                                                                                                     & perform inefficiently   when dealing with the high-dimensional space or stochastic dynamics                                      \\ \cline{2-5} 
					& ICM \cite{pathak2017curiosity}; Large Scale   Study~\cite{burda2018large}; GIRL~\cite{yu2020intrinsic}; & $r^i_t=\frac{\alpha}{2}   \|\phi({s_{t+1}})-\hat{\phi}_{s_{t+1}}\|^2_2$                                                                                                                                                                                  & improve the sample   efficiency of learning directly from raw pixels; require no $r^e_t$;   generalize well even on unseen tasks                                                                                        & $r^i_t$ may vanish   during training; may perform unstably; IDF suffers from the ``couch   potato" issue (``noisy TV" problem)   \\ \cline{2-5} 
					& Disagreement~\cite{pathak2019self};   CIExplore~\cite{yang2021ciexplore}                               & \multirow{2}{\linewidth}{$r^i_t =   \text{var}(\hat{\phi}_{s_{t+1}}(s_t, a_t))=\mathbb{E}\left[ (\hat{\phi}_{s_{t+1}}(s_t, a_t)-\phi({s_{t+1}}) )^2\right]$}                                                                                                    & stabilize the model   performance by enhancing the reliability of ICM; explore in stochastic   environments                                                                                                             & increase the   computational complexity by the ensemble amount of dynamics models; explore   extensively across all state space  \\ \cline{2-2} \cline{4-5} 
					& Plan2Explore~\cite{sekar2020planning}                                                                             &                                                                                                                                                                                                                                                          & construct the world model for zero-shot and few-shot downstream tasks; prove its relation to the information gain in VIME \cite{houthooft2016vime} & require sufficient   exploration for the world model; learn with low sample efficiency                                           \\ \cline{2-5} 
					& CCFDM~\cite{nguyen2021sample}                                                                                     & $r_t^i=\alpha \exp   (-\eta t) \text{ dissim}(q,k)\frac{r^e_{\max}}{r^i_{\max}}$                                                                                                                                                                         & improve   sample efficiency for continuous control tasks; learn a meaningful state   representation in a self-supervised way                                                                                            & may perform unstably                                                                                                             \\ \cline{2-5} 
					& PER~\cite{schaul2015prioritized}                                                                                  & $P_i =   \frac{p_i^\alpha}{\sum p_i^\alpha} $                                                                                                                                                                                                            & improve sample   efficiency through prioritized experience replay                                                                                                                                                       & cannot explicitly   improve representation learning for raw pixels; increase the computational   cost for evaluating TD errors   \\ \hline
					Complexity                                      & AMIGO~\cite{campero2020learning}                                                                                  & 
					\makecell{$r_{student}^i=1$,   $r_{teacher}^i=\begin{cases}
						a &  t^+ \geq t^* \\
						-b &  t^+ < t^*
						\end{cases}
						$}                                         & perform in procedurally-generated environments with a parameterized family of tasks;   self-propose a natural curriculum to learn a diverse range of skills                                                           & lack more abstract   forms of goals in more rich domains                                                                         \\ \hline
					Change                                          & RIDE~\cite{DBLP:conf/iclr/RaileanuR20}                                                                            & $r_t^i=\frac{||\phi(s_{t+1})-\phi(s_{t})||_2}{\sqrt{N_{ep}(s_{t+1})}}$                                                                                                                                                                                   & encourage to take the   actions with substantial influences; avoid repeatedly exploring the same   states; $r_t^i$ does not diminish to zero                                                                            & only applicable to   the tasks where agents need to influence the environments                                                   \\ \hline
				\end{tabular} }
			}
		\end{center}
		\caption{A taxonomy of the current literature on curiosity-driven RL}
		\label{fig:taxonomy_RL}
	\end{table}
\end{landscape}

\section{CDL in Recommender Systems}
\label{Recommendation}
In the domain of recommender systems, let $\mathcal{U}=\{u_1,u_2,\cdots,u_m\}$ and $\mathcal{I}=\{i_1,i_2,\cdots,i_n\}$ represent the user set and the item set with the sizes of $m=|\mathcal{U}|$ and $n=|\mathcal{I}|$, respectively. For a user–item rating matrix $\mathbf{R}_{m\times n}$, its entry $r_{i,j}$ denotes the rating from the user $i$ on the item $j$ with $r_{i,j}\in \mathbb{R}^+$ ($r_{i,j}\in \{0,1\}$ for binary class). The main objectives of conventional recommender systems are to accurately predict the user-item rating matrix $\mathbf{R}$ (individual interest) among different items and ultimately suggest the most interesting items to users (provide personalized top-$N$ ranked items). 

The traditional approaches such as content-based filtering and collaborative filtering (CF) aim to obtain high recommendation accuracy by extensively recommending items relevant to the user’s individual interest, which might cause over-fitting and over-consumption of users' potential interest. For example, latent factor models such as Matrix Factorization (MF) approximates rating $\hat{r}_{i,j}$ by the inner product of latent representations of user preferences $\mathbf{p}_i$ and item characteristics  $\mathbf{q}_j$, i.e. $\hat{r}_{i,j}=\mathbf{p}_i^T \mathbf{q}_j$. Although imposing some probabilistic prior \cite{mnih2008probabilistic} or considering temporal information \cite{koren2009collaborative} can improve the accuracy of the predicted rating, these methods tend to overly rely on users' previous access history and are incapable of encouraging richer information exploration. As a result, the recommended items do not promote diversity and users generally lack the motivation to engage in the long run. 

To resolve these issues, artificial curiosity can help to stimulate user's individual interest (curiosity) level such that it can be utilized to promote richer information discovery and sustain users' engagement in the long run. On the other hand, given additional information such as explicit social relations, the social interest inspired by a friend's recommendation can also contribute to the user's total interest. Intuitively, users become curious and interested in a particular item when their friends give unexpected or a wide range of ratings. In this section, we review the existing works of artificial curiosity that have been incorporated 1) to improve the measure of individual interest aroused by different items and 2) quantify social interest aroused from explicit social relations, where both types of curiosity-driven learning (CDL) methods ultimately improve the recommendation quality and address the overfitting issue.


\subsection{Surprisingness-Based Curiosity}
\label{recommendation:Surprisingness-Based Curiosity}
In news recommender systems, each news article may contain several topics $t$, where some topic combination such as ``basketball" and ``match" is commonly correlated, but the other combinations are quite unlikely to co-occur. Intuitively, people are more likely to be attracted by the news articles that have unexpected topic combinations, which might contain richer information and substantially arouses curiosity. Based on this motivation, \citeauthor{DBLP:conf/chiir/NiuA21}~\cite{DBLP:conf/chiir/NiuA21} proposed a health news recommender system named LuckyFind, which leverages the computational model of surprising topics. Given any health news, its surprisingness score $S$ is measured through pointwise mutual information \cite{DBLP:conf/chi/NiuAMG18} that describes how likely one topic occurs when observing the other. In particular,
\begin{equation}
S=\max_{i,j} s(t_i, t_j)=\max_{i,j} -\log_2 \frac{p(t_i,t_j)}{p(t_i)p(t_j)},
\end{equation} 
where $p(t_i,t_j)$ is the joint probability of the two topics, and $p(t_i)$ and $p(t_j)$ are respectively the individual probabilities. Based on the surprisingness score and user's preferred topics, the news articles with high $S$ will be recommended within user's preference scopes. To avoid over-presentation of non-satisfactory news, an adaptive penalized $S$ is proposed by incorporating users’ real-time negative ratings. The experimental simulation has shown that LuckyFind can reinforce users' current interests and even motivate users to explore new interests. Although it shows surprisingness can effectively arouse curiosity for long-term engagement, the model is rather naive as it cannot provide a personalized surprisingness score and it overly relies on the prior distribution of topics. In practice, the expensive prior information may not be available, so the implicit measurement by a more sophisticated model is promising.

In social relations, users have a general expectation of their friends' rating based on their own internal beliefs. As users are closely connected with their friends during social interactions, they tend to be more interested when a connected friend gives an \textit{unexpected} high rating on a particular item. To capture these responses and describe users' social interests more comprehensively, \citeauthor{wu2016social}~\cite{wu2016social} proposed a surprise-evoked curiosity-driven recommender system, where items are recommended based on both individual and social interest. Based on the curiosity framework, surprisingness $S(f,j)$ arises when user $i$’s expectation of a friend $f$’s preference towards a particular item $j$ (pseudo-predicted rating $\check{r}_{fj}$ by the MF method) is much lower than the actual rating $r_{fj}$. The surprisingness $S(f,j)$ can be expressed as
\begin{equation}
S(f,j) = \left\{
\begin{array}{ll}
r_{fj}- \check{r}_{fj}& \quad \text{if } r_{fj}- \check{r}_{fj}>T_E\\
0& \quad \text{otherwise}
\end{array}
\right. ,
\end{equation}
where $T_E$ denotes the error threshold by averaging the positive predicted rating errors. In addition to surprisingness level, the surprisingness correlation $SC(i,f)$ is also defined to describe the response strength between user $i$ and the connected friends, which can be expressed as
\begin{equation}
SC(i,f)=1-\frac{\sum_{l\in M(i,f)}\frac{|S(i,l)-S(f,l)|}{R_m}}{M(i,f)},
\end{equation}
where $M(i, f)$ represents the set of common items both surprisingly rated by $i$ and $f$, and $R_m$ is the normalization constant to ensure $SC(i,f)\in [0,1]$. Intuitively, a higher surprise correlation $SC(i,f)$ indicates that user $i$ and $f$ have given a large amount of common surprising ratings and the degree of surprisingness is more consistent. Finally, the overall social interest of user $i$ on item $j$ can be represented by the minimum, average or maximum value of $SC(i,f)S(f,j)$ for any $f$, which is weighted together with individual interest approximated by conventional MF method to provide the top-$N$ ranked items. Compared with the baseline models with only individual interest considered, this recommender system has obtained higher precision, coverage, and diversity. Nevertheless, these improvements are at the cost of the prediction accuracy and the performance might be dramatically affected by high-dimensional and sparse rating matrix/social relations. 

\subsection{Uncertainty-Based Curiosity}
Given the similar setting of social curiosity, users feel \textit{uncertain} about the item rating when their friends give inconsistent ratings on the same items. As a result of the uncertainty-based social interest (curiosity), they become more interested in the item to explore and reduce their uncertainty. In this way, the uncertainty can be utilized to measure social interest, which is elicited by the rating deviation from users' friends \cite{wu2017modeling}. In particular, the overall uncertainty of user $i$ on item $j$, $U(i,j)=(1-DS(i,j))\cdot SE(i,j)$ consists of the uncertainty about the item $SE(i,j)$ and the uncertainty about the rating distribution $DS(i,j)$. They are formulated by leveraging Shannon entropy and Dempster-Shafer theory, where the rating density $p(i,j,l)$ is constructed. Therefore, 
\begin{equation}
DS(i,j)=\frac{R}{\sum_{l=1}^{R}Count_l^i(j)+R}, \quad SE(i,j)=-\sum_{l=1}^{R}p(i,j,l)\log p(i,j,l),
\end{equation}
where  $Count_l^i(j)$ represents the count of each distinctive rating level $l$ of item $j$ given by user $i$'s friends, $p(i,j,l)$ is the probability inferred by $Count_l^i(j)$, and the rating system has a discrete rating scale with $R$ distinctive values. $DS(i,j)$ is high when the distinctive rating levels $R$ is large or only a few friends of $i$ have given the rating to the item $j$, indicating a low strength of $SE(i,j)$. Finally, the top-N recommended items are provided by weighting between individual interest approximated by the conventional MF method and rank of overall uncertainty. The proposed system achieves superior performance in precision and diversity (except for coverage), compared to the previous CDL method \cite{wu2016social}. However, it requires the prior information to be explicitly given for constructing the density approximation. Meanwhile, it remains challenging when the user-item rating matrix is sparse or not given explicitly as the uncertainty can be misleading or unable to obtain in this case.

\subsection{Novelty-Based Curiosity}
To obtain individual interest when $\mathbf{R}$ is not explicitly given, the artificial curiosity can be utilized to approximate $\mathbf{R}$ in the sense that users are more interested (high rating score) in the \textit{novel} item given the same relevance. Following this motivation, novelty-based curiosity can be integrated for measuring user's interest in the music recommender system, where novelty is quantified by a Probabilistic Curiosity Model (PCM) based on the user’s access history \cite{zhao2016much}. In the music recommender system, $\mathcal{A}$ denotes the set of artists performing the music and each item is associated with different tags $Tags$ for differentiation. To measure the degree of novelty, three important factors of \textit{frequency}, \textit{recency} and \textit{similarity} should be considered, which are inversely related to novelty and can be summarized from user's access history. 
\begin{itemize}
	\item Let $SF_{i,j}^t$ represent the novelty component measured by the scaled frequency of user $i$ accessing item $j$ before time $t$, which could be formulated as
	$
	SF_{i,j}^t=\frac{1}{2}(e^{-\rho_a\cdot |A_{i,j}^t|}+e^{-\rho_i\cdot |I_{i,j}^t|})
	$,
	where $|A_{i,j}^t|$ is the visitation times of the same artist performing $j$ accessed by $i$ before $t$, $|I_{i,j}^t|$ is the visitation times of item $j$ accessed by $i$ before $t$, and $\rho_a\in\mathbb{R}^+$ and $\rho_i\in\mathbb{R}^+$ are scaling coefficients. 
	In this formulation, when the user accesses an item and its artist more frequently, the corresponding novelty will become lower.
	\item Let $SR_{i,j}^t$ represent the novelty component measured by the scaled recency of user $i$ accessing item $j$ before time $t$, which could be formulated as 
	$
	SR_{i,j}^t=\frac{1}{2}[e^{\rho_t\cdot (t-A_{i,j}^{(-1)})}+e^{\rho_t\cdot (t-I_{i,j}^{(-1)})}]
	$
	where $I_{i,j}^{(-1)}$ and $A_{i,j}^{(-1)}$ represent the timestamp of user's last access to item $j$ and its artist, and $\rho_t\in\mathbb{R}+$ is a scaling coefficient. Intuitively, when the user accesses the item and its artist more recently, the corresponding novelty will become lower.
	\item Let $Dissim_{i,j}^t$ represent the novelty component measured by the dissimilarity between $j$ and $i$'s historically accessed items before $t$, which could be formulated as
	$
	Dissim_{i,j}^t=\frac{1}{2|Tags(j)|}\sum_{tag}^{Tags(j)} (e^{-\rho_{tag}\cdot |I_{i,tag}|}+e^{\rho_t\cdot (t-I_{i,tag}^{(-1)})})
	$,
	where $|Tags(j)|$ is number of tags associated with $j$, $|I_{i,tag}|$ is the visitation times of the same $tag$ by $i$, $I_{i,tag}^{(-1)}$ is the timestamp of user's last access to items with $tag$, and $\rho_{tag}\in\mathbb{R}+$ is a scaling coefficient.  
	Intuitively, the more common tags the evaluated item has with historical items, the less novel the evaluated item will be.
\end{itemize}
Hence, the overall novelty $\text{Novelty}_{i,j}^t$ can be obtained by averaging the above three factors:
$
\text{Novelty}_{i,j}^t=\frac{1}{3}(SF_{i,j}^t+SR_{i,j}^t+Dissim_{i,j}^t)
$. An item (music) is more novel when the user accesses it and its artist less frequently and less recently, or when an item has fewer common tags.
As the aroused curiosity follows an inverted U-shaped distribution w.r.t. novelty, the curiosity function is estimated by beta distribution based on the measured novelty $ \hat{C}_{i,j}=Beta(\text{Novelty}_{i,j}; \hat{\alpha}_i,\hat{\beta}_i)$. Substantially, the Top-N item recommendation list could be obtained through joint optimization of relevancy and curiosity or the MF method. By comparing with Popularity, Item-CF and Ranking-MF, the proposed curiosity-based recommender system could significantly outperform in terms of inter user similarity and novelty fitness despite some sacrifice in recommendation precision. Remarkably, it achieves the personalized curiosity estimation with the consideration of item novelty to enhance the recommendation accuracy. Given the context of music recommendations, it is possible to include previously accessed items into the recommendation list. In addition to the trade-off of precision, it remains to validate the generalization ability in other domains and investigate the possibility of incorporating other collative variables.

\subsection{Novelty-and-Conflict Based Curiosity}
In many practical scenarios such as e-commerce, it is expensive or infeasible to acquire a large amount of training data prior to learning. Instead, training data arrives in a sequential manner through users' engagement. In the early stage with only limited samples, individual interest derived by traditional methods becomes unreliable due to overfitting. To remedy this, \citeauthor{xu2019enhancing}~\cite{xu2019enhancing} proposed a recommender system with a Stimulus-evoked Curiosity mechanism (SeCM). 
In SeCM, the stimulus intensity is evaluated by weighting novelty and conflict. In particular, novelty is measured by prior ratings on other items in chronological order and the dissimilarity between items. $
\text{Novelty}(i,j)=\sum_{l\in \mathcal{I}_i^{t_{max}}(j)}(e^{-\mu\cdot \text{Position}(l)}\cdot s_{jl})
$,
where $\mathcal{I}_i^{t_{max}}(j)\subset \mathcal{I}$ denote the set of the latest $t_{max}$ items that user $i$ has rated prior to item $j$, $\text{Position}(l)$ denotes the position of item $l$ in $\mathcal{I}_i^{t_{max}}(j)$ arranged in a chronological order, the exponential term denotes the memory decaying process by time, $\mu$ denotes a constant coefficient, and $s_{jl}$ denotes the re-scaled Pearson Correlation Coefficient (PCC) (dissimilarity between item $i$ and $l$). Thus, it can be easily proved that more recent ratings and larger dissimilarity $s_{jl}$ can increase the novelty of the evaluated item.

On the other hand, users encounter high conflict when they receive both positive ratings and negative ratings from close friends. The conflict of user $i$ on item $j$ can be measured based on the nearness to equality in the competing strengths between negative and positive ratings for the same item $j$ given by $i$'s friends. Let $\mathcal{U}_{i,j}^{+}$ and $\mathcal{U}_{i,j}^{-}$ denote the set of $i$'s friends who gave ratings higher and lower than $\bar{r}$ in the descending order for item $j$, respectively. Then, the strength of positive ratings $\text{Pos}_{ij}$ and negative ratings $\text{Neg}_{ij}$ on item $j$ provided by $i$'s friends is defined as
\begin{equation}
\text{Pos}_{ij}=\frac{\sum_{f\in \mathcal{U}_{i,j}^{+}}(PCC_{if}\cdot(r_{f,j}-\bar{r}))}{\sum_{f\in \mathcal{U}_{i,j}^{+}}PCC_{if}}, \quad \text{Neg}_{ij}=\frac{\sum_{f\in \mathcal{U}_{i,j}^{-}}(PCC_{if}\cdot(\bar{r}-r_{f,j}))}{\sum_{f\in \mathcal{U}_{i,j}^{-}}PCC_{if}}.
\end{equation}
Therefore, the conflict $\text{Conflict}(i,j)\in[0,1]$ can be ultimately computed by 
\begin{align}
\text{Conflict}(i,j)=1-\left|\frac{\text{Pos}_{ij}-\text{Neg}_{ij}}{\text{Pos}_{ij}+\text{Neg}_{ij}}\right|
\end{align}
where a higher value close to 1 represents a greater conflict when both positive strength and negative strength are strong, indicating a higher incompatibility of the two types of ratings on $j$ given by $i$’s social connections.

By integrating the measured novelty and conflict, the overall item intensity $\text{Int}$ for user $i$ on item $j$ can be expressed as
$
\text{Int}(i,j)=\alpha \text{Novelty}(i,j)+(1-\alpha)\text{Conflict}(i,j)
$, where $\alpha$ is a coefficient for weighting. With the overall intensity, the aroused curiosity level $\hat{C}_i(\text{Int})$ can be approximated by aggregating a reward system $\text{Reward}_i(\text{Int})$  and an anxiety aversion $\text{Anxiety}_i(\text{Int})$ as illustrated in Figure \ref{InvertedU}, with the consideration of individual preferences.
For each user $i$, the Stimulus-evoked Curiosity mechanism (SeCM) can be formulated as
\begin{equation}
\text{Reward}_i(\text{Int})=\frac{1}{1+e^{-\theta_{\text{Reward}}(\text{Int}-\text{Int}_i^{\text{Reward}})}}, \quad
\text{Anxiety}_i(\text{Int})=\frac{1}{1+e^{\theta_{\text{Anxiety}}(\text{Int}-\text{Int}_i^{\text{Anxiety}})}},
\end{equation}
where $\theta_{Re}$ and $\theta_{Pu}$ represent constant coefficients, and $\text{Int}_i^{\text{Reward}}$ and $\text{Int}_i^{\text{Anxiety}}$ represent two individual preference thresholds that needs to be optimized for each user $i$. 

Finally, this recommender system provides the top-N recommended items based on user preference computed by weighting-based MF and user curiosity $\hat{C}_i(\text{Int}(i,j))$. The empirical experiments on various datasets indicate that the proposed curiosity-driven recommender system could approximate individual curiosity preference consistent with Figure \ref{InvertedU} and help to diversify the recommended items. However, it sacrifices the accuracy and the model is incapable of handling the dataset with sparse ratings and social relations. Compared to PCM \cite{zhao2016much} that utilized novelty-based curiosity, the novelty in SeCM is not comprehensively measured and can be misrepresented. Furthermore, it remains to further incorporate more collative variables with the consideration of personalized weights.

\subsection{Discussion}

Curiosity-driven recommender system is a relatively new research area, where only a few works have been carried out to incorporate curiosity mechanisms into recommendation tasks. Traditional recommender systems aimed to achieve high recommendation accuracy by extensively recommending items relevant to the user’s past access history, which might cause over-fitting problems and lose users' interest towards the recommended items for long-term engagement. Besides the models mentioned above, some researchers tried to model user's curiosity explicitly based on questionnaires and surveys \cite{menk2015hybrid}, but it is beyond the scope of this paper. With some acceptable expense of accuracy, curiosity-driven recommender system can outperform traditional methods in terms of \textit{coverage}, \textit{diversity} and \textit{personalization}, which are more desirable in real-world applications.

However, some improvement can be carried out and further research should focus on the following aspects:
\begin{itemize}
	\item By exploiting the social network curiously, a curiosity-driven recommender system should be capable of generalizing to unrated items or new users.
	\item Besides various MF methods, there are other types of techniques such as collaborative deep learning \cite{wang2015collaborative} and collaborative topic regression \cite{wang2011collaborative,purushotham2012collaborative} performing better than MF. It remains to investigate how to incorporate CDL mechanism into these models to resolve more challenging recommendation tasks. For example, \citeauthor{han2019curiosity}~\cite{han2019curiosity} proposed a curiosity-driven recommender system for personalized learning by leveraging the deep RL technique; however, it still suffered from the ``couch potato" issue.
	\item The existing curiosity-driven recommender systems only utilize one or two collative variables to model curiosity. To obtain robust performances in different recommendation tasks, a principled system with more collative variables and more sophisticated mechanisms is preferred.
	\item When modeling the personalized curiosity distribution, it is of great significance to explore Bayesian models with prior knowledge in probabilistic curiosity models.
	\item The robustness of curiosity-driven recommender systems needs to be validated and further improved when dealing with high-dimensional and sparse rating data. Moreover, most social recommendations with CDL largely depend on the explicit form of social relations, and the investigation on the implicit social relation exploitation is currently missing. 
\end{itemize}

\section{CDL in Classification}
\label{classification}
In the domain of supervised classification, given input features $\mathbf{X}$, the ultimate goal is to find a mapping function $\mathcal{M}: \mathcal{X} \mapsto \mathcal{Y}$ to predict the class label $\mathbf{Y}$ accurately. In the online-sequential setting, training data are collected in a one-by-one or batch-by-batch manner, so the classier needs to be updated in a sequential manner. During training, overfitting issues may occur due to limited samples and the performance is often unsatisfying due to poor generalization. Meanwhile, it requires an adaptive way to optimize model complexity, thereby reducing the computational cost. As curiosity can guide agents to actively explore and learn with limited prior knowledge as shown in previous sections, it is a natural choice to incorporate artificial curiosity in the classification task to address the above issues. 

In general, entropy-based classifiers and their variants can be approximately viewed as leveraging the uncertain-based curiosity, where their main objective is to reduce the uncertainty of labeling the given samples with different constraints. Similarly, the classifiers with the main objective of minimizing mean squared error can be approximately regarded to leverage the surprisingness-and-incongruity-based curiosity. While these classifiers are within the scope of curiosity-driven learning, they have been exhaustively reviewed and compared in the many survey works \cite{kotsiantis2007supervised,gaber2007survey,huang2011extreme,sen2020supervised}. As our primary interest in this paper is to review the curiosity-driven learning (CDL) methods that significantly differ from traditional approaches, we would like to focus on one particular work in this section, which integrates the artificial curiosity by multiple collative variables.

\subsection{Curiosity-Driven Learning Strategy}
In the traditional method for online learning, Extreme Learning Machine (ELM) \cite{6035797} utilizes a single-layer feedforward neural network (SLFN) with $k$ additive or radial basis function (RBF) hidden neurons $G$, $
\mathbf{Y}=\mathbf{H} \pmb{\beta}
$
where $\mathbf{H}=\begin{bmatrix}
G(\mathbf{a}_1,b_1,\mathbf{x}_1) & \cdots & G(\mathbf{a}_k,b_k,\mathbf{x}_1)\\
\vdots & \cdots & \vdots\\
G(\mathbf{a}_1,b_1,\mathbf{x}_n) & \cdots & G(\mathbf{a}_k,b_k,\mathbf{x}_n)
\end{bmatrix}_{n \times k}
$, 
$\pmb{\beta} \in \mathbb{R}^{k\times l}
$, $n$ is the sample size, and $l$ is the total number of classes. In particular, ELM has a universal approximation capability and low computational cost \cite{6035797}. The hidden weights $\pmb{\beta}$ are determined iteratively, but the neuron number has to be carefully pre-selected with randomness effect \cite{liang2006fast}. Aiming to avoid manually tuning of hidden neuron numbers and facilitate the classifier to add and delete hidden neurons, \citeauthor{wu2015c}~\cite{wu2015c} proposed a Curious ELM (C-ELM), which is the first attempt to introduce the curiosity mechanism into online classifiers. 

In C-ELM, four collative variables are introduced to trigger different learning strategies. \textit{Novelty} is measured by the degree of dissimilarity (spherical potential) between the input data and the network's current knowledge. \textit{Uncertainty} is measured by the posterior probability of the prediction with the hinge-loss error. \textit{Conflict} is defined by the competing strengths between the most two possible classes. \textit{Surprisingness} is quantified by the prediction error between the predicted class and the true class. 
\begin{itemize}
	\item Novelty $\mathcal{N}(\mathbf{x})$ is measured by the degree of dissimilarity between the input data and the network's current knowledge, where the similarity is measured by the spherical potential $
	\frac{1}{k}\sum_{i=1}^{k}G(\mathbf{a}_i,b_i,\mathbf{x})$. Therefore, the novelty of the new input data $\mathbf{\left\{x\right\}}_{1\times m}$ with size of $1$ and features of $m$ can be quantified by  
	\begin{equation}
	\mathcal{N}(\mathbf{x})=1-\frac{1}{k}\sum_{i=1}^{k}G(\mathbf{a}_i,b_i,\mathbf{x})
	=1-\frac{1}{k}\sum_{i=1}^{k}g(b_i\begin{Vmatrix}
	\mathbf{x}-\mathbf{a}_i
	\end{Vmatrix}),
	\end{equation}
	where $k$ represents the total number of neurons in SLFN at the current stage. $\mathcal{N}(\mathbf{x}) \in \left[0,1\right]$. Intuitively, a higher value of $\mathcal{N}(\mathbf{x})$ represents that the input data is more novel since it is different to a higher degree according to network's current knowledge.
	
	\item Uncertainty $\mathcal{U}(\mathbf{x})$ is measured by how unconfident the C-ELM classifier is in its estimated class. In particular, the confidence can be measured by the posterior probability of the prediction with hinge-loss error $\mathbf{e}$. Let $c$ represent the true class of the input data and $\hat{c}$ represent the estimated class by the classifier with $c,\hat{c}\in\{1,2,3,...,l\}$ (Note: $\hat{c}=\operatorname*{argmax}_j y_j, \quad j=1,\cdots,l $). The hinge-loss error of the estimated class $\hat{c}$ is 
	$$
	e_{\hat{c}}= \left\{
	\begin{array}{ll}
	0 & \quad \text{if } y_{\hat{c}}\hat{y}_{\hat{c}} > 1\\
	\min (\max(\hat{y}_{\hat{c}}-y_{\hat{c}},-1),1) & \quad \text{otherwise}
	\end{array}
	\right.
	$$
	Therefore, the posterior probability of input $\mathbf{x}$ belonging to class $\hat{c}$ is defined as:
	$$
	p(\hat{c}\mid\mathbf{x})=\frac{e_{\hat{c}}+1}{2}
	$$
	Substantially, the uncertainty $\mathcal{U}(\mathbf{x})$ of the predicted class $\hat{c}$ is 
	$$
	\mathcal{U}(\mathbf{x})=1-p(\hat{c}\mid\mathbf{x}), \quad \mathcal{U}(\mathbf{x}) \in [0,1]
	$$  
	\item Conflict $\mathcal{C}(\mathbf{x})$ is measured by the competing strengths of the most and the second most possible classes. Let $\hat{c}_1$ and $\hat{c}_2$ represent the most and the second most possible classes by the C-ELM classifier, i.e. $\hat{y}_{\hat{c_1}}$ and $\hat{y}_{\hat{c_2}}$ are the largest and second largest values among all $\hat{y}_j$, respectively. Thus,  the conflict $\mathcal{C}(\mathbf{x})$ of the given input $\mathbf{x}$ is defined as
	$$
	\mathcal{C}(\mathbf{x}) = \left\{
	\begin{array}{ll}
	1-\frac{|\hat{y}_{\hat{c_1}}-\hat{y}_{\hat{c_2}}|}{|\hat{y}_{\hat{c_1}}+\hat{y}_{\hat{c_2}}|} & \quad \text{if } \hat{y}_{\hat{c_1}}\hat{y}_{\hat{c_2}}> 0\\
	0 & \quad \text{otherwise}
	\end{array}
	\right.
	$$
	Note that $\mathcal{C}(\mathbf{x}) \in [0,1]$, where a larger value of $\mathcal{C}(\mathbf{x})$ implies a higher conflict between the most possible classes.
	
	\item Surprisingness $\mathcal{S}(\mathbf{x})$ occurs when the predicted class $\hat{c}$ differs from the true class $c$, which can be measured by the prediction errors considering both the predicted class $e_{\hat{c}}$ and the true class $e_{{c}}$. Thus, the surprisingness $\mathcal{S}(\mathbf{x})$ of the given input $\mathbf{x}$ can be quantified by
	$$
	\mathcal{S}(\mathbf{x}) = \left\{
	\begin{array}{ll}
	|e_{\hat{c}}e_{c}| & \quad \text{if }\hat{c}\neq c \\
	0 & \quad \text{otherwise}
	\end{array}
	\right.
	$$
	Note that $\mathcal{S}(\mathbf{x}) \in [0,1]$, where a larger value of $\mathcal{S}(\mathbf{x})$ indicates the a larger deviation between the predicted class and true class.
\end{itemize}
To enable the classifier to self-supervise its learning process, three rule-based learning strategies (specific exploratory behaviors) are proposed according to different curiosity levels, considering the measured four collative variables simultaneously.
\begin{itemize}
	\item \textbf{Neuron Addition Strategy}: When a misclassification (high surprisingness) is caused by the newness of knowledge (high novelty) with high uncertainty in its prediction, the classifier should add a new neuron to capture the new knowledge. In other words, when $\mathcal{N}(\mathbf{x}) > \theta_{\mathcal{N}^{\text{add}}}$ AND $\mathcal{U}(\mathbf{x}) > \theta_{\mathcal{U}}$ AND $\mathcal{S}(\mathbf{x}) > \theta_{\mathcal{S}}$, a new neuron needs to be added in the SLFN ($k$ will be increased by 1) with $\mathbf{a}_{\text{new}}=\mathbf{x}_{\text{new}}$ and $b_{\text{new}}$ assigned with random value, where  $\theta_{\mathcal{N}^{\text{add}}}, \theta_{\mathcal{U}}$ and $\theta_{\mathcal{S}}$ are the neutron addition thresholds for novelty, uncertainty and surprisingness, respectively. Meanwhile, the class label of the newly added neuron will take the true class label of the input data, i.e. $L_{k+1}=c$. Hence, $\mathbf{H}$ will be appended with $G(\mathbf{a}_{\text{new}},b_{\text{new}},\mathbf{x})$ and $\pmb{\beta}$ will be recalculated.
	
	\item \textbf{Neuron Deletion Strategy}: When a misclassification (high surprisingness) happens on a familiar stimulus (low novelty) caused by high competing strengths between two decisions (high conflict), the classifier should adjust its decision-making rules by deleting the neuron most contributing to the incorrect class. Substantially, the correct decision is strengthened and the incorrect decision is weakened. In other words, when $\mathcal{N}(\mathbf{x}) < \theta_{\mathcal{N}^{\text{del}}}$ AND $\mathcal{C}(\mathbf{x}) > \theta_{\mathcal{C}}$ AND $\mathcal{S}(\mathbf{x}) > \theta_{\mathcal{S}}$, the classifier should delete the most improper hidden neuron
	directed towards the incorrect predicted class: $\left\{\text{del} \mid \max{G(\mathbf{a}_{\text{del} },b_{\text{del} },\mathbf{x})} \text{ AND } L_{\text{del}}=\hat{c}\right\}$, where $\theta_{\mathcal{N}^{\text{del}}}, \theta_{\mathcal{C}}$ and $\theta_{\mathcal{S}}$ represent the neutron deletion thresholds for novelty, conflict and surprisingness, respectively. Subsequently, the selected neuron will be removed from the SLFN ($k$ will be decreased by 1) and $G(\mathbf{a}_{\text{del}},b_{\text{del}},\mathbf{x})$ shall be removed from $\mathbf{H}$ as well.	
	
	\item \textbf{Parameter Update Strategy}: If neither the neuron addition strategy nor the neuron deletion strategy is triggered, the input data shall be considered as the familiar knowledge to the classifier. Therefore, it only needs to update the output weight $\pmb{\beta}$ analytically.
\end{itemize}



\subsection{Discussion}
In this section, a most representative CDL algorithm named C-ELM for the online classification task is reviewed. In particular, C-ELM consists of two main components: an internal cognitive component with an SLFN and a curiosity arousal component containing curiosity-based strategies to regulate the learning process. Moreover, it has comparable advantages over the traditional methods. Firstly, it avoids manually setting the number of hidden neurons $k$, which is usually determined by trial-and-error. Secondly, it has a flexible evolving network structure by enabling the classifier itself to add and delete hidden neurons dynamically to determine the optimal structure. Thirdly, by assigning the parameter of RBF centers $\mathbf{a}_i$ instead of randomization, the high randomization effect in the prediction results could be reduced significantly.

Since curiosity has an inherently motivated learning mechanism, C-ELM performs well in various datasets without compromising the merits of traditional ELM. Its generalization performance is better than other classifiers such as SVM \cite{wang2005support}, ELM \cite{6035797} and McELM \cite{savitha2014meta} with less computational time, fewer hidden neurons and higher prediction accuracy. However, the success largely depends on the proper tuning of the thresholds for neuron addition/deletion strategies. Moreover, the direct manipulation of neurons can even dampen the stability of SLFN. During model training, the incoming data has to be trained in a one-by-one manner, which is quite inefficient. As C-ELM is the only work to incorporate a sophisticated curiosity mechanism in online classification tasks, it is also interesting to address other challenging issues such as over-fitting, high-dimensional data, imbalanced data as well as model sparsity in future research.

\section{Conclusion and Future Research}
\label{conclusion}
In this paper, we have comprehensively reviewed the work of merging curiosity mechanism and AI techniques. In particular, the theories and mechanism of psychological curiosity are introduced and a unified framework is presented for formulating artificial curiosity in a computational way. Subsequently, the recent works on curiosity-driven learning (CDL) in Reinforcement Learning, Recommendation, and Classification are reviewed, which have demonstrated comparable learning performances and desirable exploratory behaviors with self-organized learning. 

The current AI techniques tend to exhaustively mine patterns from abundant data, while most of them fail to demonstrate the full and authentic functionality of human beings. More specifically, most of the methods are incapable of learning continuously and dynamically to update their learning rules, and the reasoning of these methods does not reveal how knowledge is represented and processed in a human-like way. To further bridge the gap between machine-based intelligence and human-based intelligence, future CDL research can attempt to incorporate curiosity mechanisms into the domains of active learning, few-shot learning, meta-learning, and explainable AI, where curiosity can be exploited to boost learning and efficiency. In particular, enabling AI agents to self-identify the interesting patterns with limited training samples while generalizing well in different problem settings are the most desirable features in these domains.

One issue in the existing CDL techniques is that they generally fail to consider the interaction and correlation between collative variables. In fact, the aroused curiosity is a complex process as the result of the interaction and integration of different collative variables. For instance, uncertainty and surprisingness may emerge together when novelty arises. Moreover, performance evaluation of CDL methods is another challenge. While the prediction performance can be assessed by traditional ML metrics, there is no suitable benchmark for validating the model performance in terms of stimulated human curiosity. It requires cross-disciplinary expertise in psychology and neuroscience. Therefore, it will also be an interesting topic in future research to address the feasibility of cognitive validation. 

In addition, there may be some other potential concerns raised up when leveraging CDL techniques, such as risk management and privacy issues, where agents may perform risky actions or arbitrarily mine sensitive information that could cause potential danger or discomfort.
In psychology, it has been found that excessively seeking curiosity might cause distractibility, which is a symptom of attention-deficit or hyperactivity disorder \cite{silver2008attention}. In the expression “curiosity killed the cat”, curiosity is often considered dangerous when this noble drive grows in an unrestricted manner. Similarly, artificial curiosity encourages agents to understand and control in a fully sufficient and rational way by collecting even personal information and disclosing the patterns that could harm to individuals. Therefore, it is of great significance to build a clear boundary of safety and privacy for curiosity-driven exploration such that agents will be capable of filling the information gap under proper regulations.

As many real-world tasks require both precision and efficiency, CDL is a natural choice to harness the high precision of state-of-the-art techniques and the high efficiency in driving cognitive development by inferring agents' psychological states. For example, an autonomous driving system should anticipate a child suddenly running across the road due to his/her curious psychology towards a toy shop on the other side. In this application, it is crucial for the agent to model and stimulate curiosity based on the observations. As for companion robots, they are expected to personalize interactions with a diverse range of users by maintaining their cognitive-affective (curiosity) states in the range of excitement instead of boredom or confusion. Therefore, we believe curiosity-driven learning can play an important role in the next-generation artificial intelligence, which seamlessly facilitates AI agents to explore, learn and generalize.

\bibliographystyle{ACM-Reference-Format}
\bibliography{sample-base}

\appendix
\end{document}